\documentclass[letterpaper]{article} % DO NOT CHANGE THIS
\usepackage[preprint]{aaai2027}  % Public preprint: show authors without a proceedings copyright notice
\usepackage[hyphens]{url}  % DO NOT CHANGE THIS
\usepackage{graphicx} % DO NOT CHANGE THIS
\usepackage{natbib}  % DO NOT CHANGE THIS AND DO NOT ADD ANY OPTIONS TO IT
\usepackage{caption} % DO NOT CHANGE THIS AND DO NOT ADD ANY OPTIONS TO IT
\usepackage{amsmath}
\usepackage{amssymb}
\usepackage{booktabs}
\usepackage{tabularx}
\usepackage{array}
\usepackage{makecell}
\usepackage{multirow}
\usepackage{algorithm}
\usepackage{algpseudocode}
\usepackage{placeins}
\usepackage{flafter}
\usepackage{dblfloatfix}
\usepackage{adjustbox}

\usepackage[most]{tcolorbox}
\usepackage{listings}

\title{Tree-of-Experience: Hierarchical Experience Management for Self-Evolving Agents}
\author{
Zihao Deng, Yining Zhu, Leiming Wang, Junbo Wang, Jingfei Lu
}
\affiliations{}

\begin{document}

\maketitle

%平铺直叙，重写
\begin{abstract}
Continual self-evolution requires LLM agents to transform environmental interactions into reliable and reusable experience. Existing methods typically refine individual trajectories or abstract shared knowledge from related trajectories, but their experience representations are often disconnected from the underlying reasoning process. This limits feedback attribution, cross-task transfer, and update and retrieval efficiency, particularly in complex reasoning tasks with outcome-level feedback. To overcome this limitation, we propose \textbf{T}ree-\textbf{o}f-\textbf{E}xperience (ToE), a structured experience-management framework that aligns experience organization with the hierarchical reasoning process of LLM agents. Specifically,
ToE organizes the experience into a shared tree of analytical perspectives and reasoning paths, whose reliability is calibrated through environmental outcomes to support systematic updating, transfer, and efficient retrieval. The experimental results on \textsc{Game of 24} and \textsc{FinEvolveBench} show that ToE substantially improves both problem-solving performance and efficiency. On \textsc{Game of 24}, ToE achieves a 31.4\% relative improvement in accuracy over the experience-free ToT baseline. On \textsc{FinEvolveBench}, ToE improves tsIC by an average of 41.24\% over the experience-free pipeline across 12 evaluation settings, whereas conventional experience-management methods often underperform experience-free baselines.

% ToE integrates experience across task instances into a shared tree, where nodes represent analytical perspectives and root-to-leaf paths represent lines of reasoning. Environmental outcomes are attributed to the corresponding reasoning components to calibrate their reliability, enabling systematic updates, perspective-level transfer, and efficient retrieval. 
% We evaluate ToE on \textsc{Game of 24}, a controlled multi-step reasoning task, and \textsc{FinEvolveBench}, a low-repetition benchmark with delayed and implicit outcome feedback. Existing experience-management methods can degrade performance when outcome feedback cannot be reliably attributed, whereas ToE consistently improves problem solving. On \textsc{Game of 24}, ToE improves accuracy by 20.4 percentage points while reducing LLM calls by 78.7\%. On \textsc{FinEvolveBench}, it achieves an average \textbf{tsIC} improvement of 41.24\% across all settings and remains effective across different backbone LLMs. These results demonstrate the importance of reasoning-aligned experience organization for complex problem solving and continual agent evolution.
\end{abstract}
\section{Introduction}
\label{sec:introduction}
Advances in reasoning, tool use, and memory have extended the use of large language models (LLMs) from autoregressive text generators to general-purpose problem-solving agents.
Yet a truly general-purpose problem solver should be able to gain experience through continual interaction with the environment and adaptively self-evolve. The key challenge is that interaction history does not automatically become experience, as experience is not merely a verbatim record of the past but a structured interpretation of what mattered, why it mattered, and when it may matter again. Only by reflecting on past interactions, filtering incidental details, and extracting transferable experiences can an agent use prior outcomes to solve related problems more effectively.

\begin{figure*}[!t]
\centering
\includegraphics[width=\textwidth]{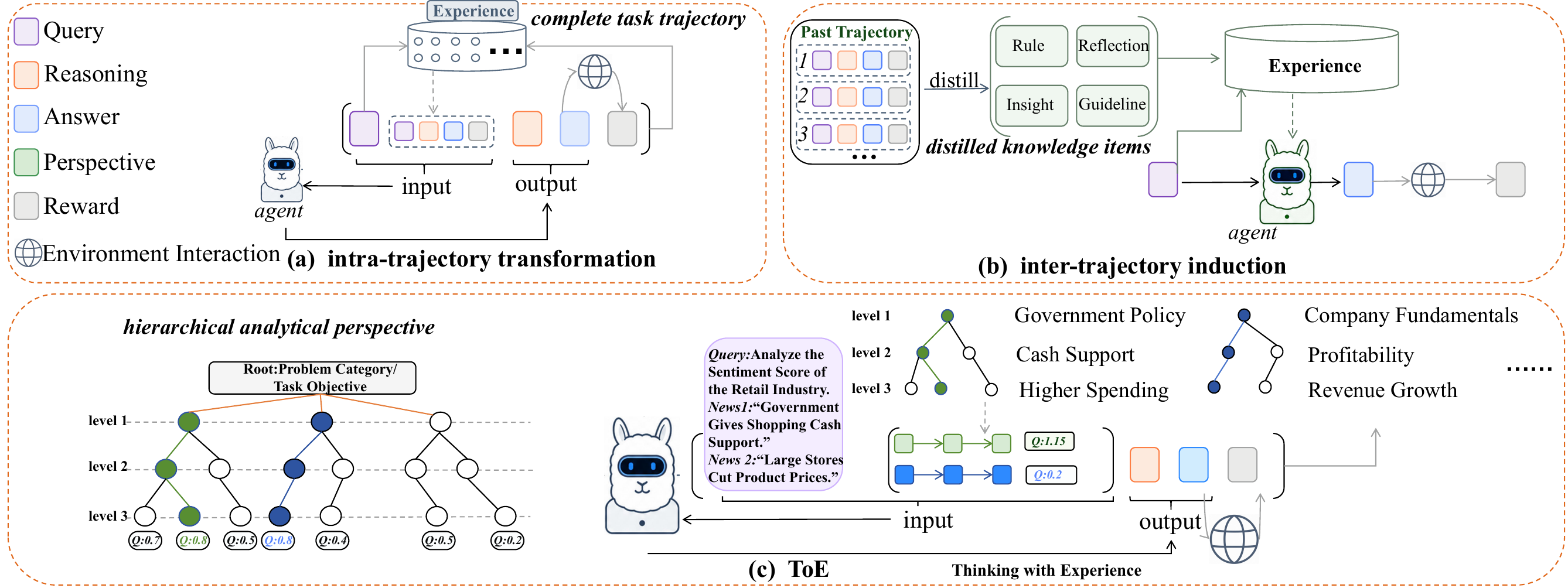}
\caption{
Comparison of experience-management paradigms.
(a) Intra-trajectory transformation stores or transforms individual task trajectories;
(b) inter-trajectory induction distills reusable knowledge from multiple past trajectories;
and (c) Tree-of-Experience (ToE) organizes experience as hierarchical analytical perspectives and retrieves a structured reasoning path for inference.
}
\label{fig:diffTask}
\end{figure*}

% 目前有两大类方法，一种是Intra-trajectory transformation，将Refined memory unit tied to the original task
% context，然后Retrieved to assist semantically similar past tasks at
% inference time。代表性的工作有Reflexion (Shinn et al., 2023), CLIN (Majumderet al., 2023), AgentFold (Ye et al., 2025)。这类方法的局限在于经验的可复用性差，Although reflection effectively mitigates noise and
% hallucinations, reflected memories are frequently
% fragmented and exhibit a high degree of depen -
% dence on context.This results in significant costs
% for retrieval and a heavy burden of inference for
% memory mechanisms when addressing new tasks.

Prior studies have explored various experience abstraction mechanisms that transform raw interaction trajectories into compact and useful experiences. These methods fall into two categories: intra-trajectory transformation and inter-trajectory induction.
Intra-trajectory transformation Reflexion\citep{shinn2023reflexion},MemRL \citep{memrl2026} and ReMe \citep{cao2026remember} refines each trajectory into experience units that remain tied to its original task context and are retrieved at inference time to assist semantically similar tasks.
Although such methods reduce noise and hallucinations in raw trajectories, the resulting memories are often fragmented and context-dependent, limiting their transferability while increasing retrieval costs and inference overhead on new tasks.

% 另一类是通过Inter-trajectory induction，对一组相关的 trajectories进行归纳总结，形成可复用的经验。Applicable to unseen scenarios as a policy prior,
% without trajectory-level matching，代表性工作有FLEX (Cai et al., 2025b), MemSkill (Zhang et al., 2026), SkillRL (Xia et al., 2026)。这种方法compresses the originally vast repository of mem-
% ory and enables generalization to unknown environ-
% ments through a form of intuition similar to that of
% humans. 
Inter-trajectory induction such as FLEX \citep{cai2025flex}, SkillGen \citep{ding2026skillgen}, and SkillRL \citep{xia2026skillrl} summarizes patterns across related trajectories to derive reusable experience, which can be served as a policy prior for unseen scenarios without relying on trajectory-level matching. By compressing large trajectory repositories into generalized experiences, these methods improve generalization to unseen environments. Their effectiveness, therefore, depends critically on the abstraction mechanism used to transform groups of raw trajectories into broadly applicable experience.

% 在这类方法中，一个好的mechanism for
% abstraction serves as the core operator for the trans-
% formation of groups of raw interaction trajectories
% into experience that is universal至关重要。
% 我们认为一个好的 abstraction mechanism 应该具有以下几个特点：
% 1. Correctness / Fidelity：identify decision-relevant factors and their associations with observed outcomes
% 2. Reusability / Transferability：一个好的经验应该适当的抽象，提炼出可以迁移到其他相关任务中的分析原则、判断标准或推理模式。abstraction at the shallow level
% retains a portion of semantic logic, utilizing “rules”
% described in natural language as experience (Cao
% et al., 2025; Chen et al., 2025d; Wei et al., 2025a;
% Hayashi et al., 2025); abstraction at the intermedi-
% ate level completely removes redundancies of natu-
% ral language, extracting only modular skeletons for
% execution as experience (Wang et al., 2024d; Liu
% et al., 2025g; Yu et al., 2025a); and abstraction at
% the deep level compresses the distribution of tra-
% jectories into the weights of the model, enabling
% the complete transformation of experience into in-
% tuition for decision making (Cheng et al., 2025b;
% Luo et al., 2025b; Wang et al., 2025b).（这里需要简写，去掉引用）
% 3. 可演化性（Evolvability）：经验不是一次生成后永久不变的规则。随着新反馈到达或环境变化，好的经验表示应支持：
% 强化已有经验；
% 修正部分内容；
% 合并重复经验；
% 拆分过于宽泛的经验；
% 弱化已经失效的经验；
% 删除长期不可靠的经验。
% 4. 可组合性/易调用性：复杂任务通常不是依赖单条经验解决的，而需要组合多个互补的分析角度。能够在inference的时候高效的调取到最有利于大模型推理的数条相关经验。
% 5. Efficiency / Compactness：经验总结应具有较高的信息密度，能够被高效的取用。
We identify four essential properties of an effective experience-abstraction mechanism: (1) \textit{Attributability}: it should attribute observed outcomes to outcome-relevant factors while filtering out incidental and task-specific details; (2) \textit{Transferability}: it should extract principles or reasoning patterns that generalize across related tasks; (3) \textit{Evolvability}: it should support the efficient accumulation, calibration, consolidation, and removal of experience as new feedback arrives or the environment changes; and (4) \textit{Efficiency}: it should organize, retrieve, and apply experience with limited storage, computational, and contextual overhead.

 % 我们认为现有工作在这几个维度，均有一些不足。例如在Correctness方面，一些abstract的方法仅总结相关trajactories的经验，对环境outcome缺乏全面的反馈分析；在Efficiency 方面，有些工作在inference的时候总结经验，十分的低效；这些我们将在related work中具体分析。

  %也可以把这四个方面都细说一下。或者We will provide a more detailed analysis of these limitations in the related-work section.

% To address these limitations, we propose \textbf{Tree of Experience} (ToE), a structured experience-management paradigm that focusing on solving complex problems
% . Instead of inducting experience from a set of similar trajectories, ToE 从解决问题的方式出发，从大模型推理的原理出发，去构建结构化思维的experience，因为经验最终会作为上下文帮助大模型进行更好的reasoning，那我们直接在构建experience的时候，构建大模型思维树，并通过与环境的interaction去量化每一条思维路径的可行性和正确性。
% manage experience from a structured thinking perspective. ToE abstracts the key analytical perspectives involved in problem solving and organizes experience into 有层级的树结构，where structured units that can be independently retrieved, updated, and composed. This formulation not only reduces the cost of experience storage and retrieval, but also improves the efficiency with which language-model agents reuse prior knowledge and continually refine their problem-solving capabilities.
However, existing methods fall short along one or more of these dimensions, particularly in low-repetition tasks where outcome-level feedback is uncertain and may arise from multiple interacting factors. Under such conditions, conventional approaches struggle to attribute outcomes to specific reasoning components and induce reliable experience.

To address these limitations, we propose \textbf{T}ree-\textbf{o}f-\textbf{E}xperience (ToE), a structured experience-management paradigm designed specially for complex problem solving. Instead of inducing experience solely from sets of similar trajectories, ToE organizes experience according to the reasoning structure of the problem itself. Since experience ultimately serves as context for future reasoning, ToE represents key analytical perspectives as a hierarchical tree and uses environmental outcomes to evaluate and update the corresponding reasoning paths. Its experience units can be independently retrieved, revised, and composed, providing improved feedback attribution, transferability, evolvability, and efficiency.

\textbf{Attributability.} As illustrated in Fig.~\ref{fig:diffTask}, compared with intra-trajectory transformation and  inter-trajectory induction, ToE integrates reasoning processes from different task instances into a shared experience tree. The root represents the problem category or task objective, while each non-root node represents a critical reasoning perspective. Parent nodes capture general analytical dimensions, whereas child nodes refine, extend, or diversify them into more specific reasoning directions. A reasoning process is therefore represented as a progressive path through the tree.

Because the experience structure mirrors the reasoning structure, an observed outcome can be propagated to the specific perspectives involved in the decision. This enables localized reinforcement or revision of relevant experience units. In contrast, conventional inter-trajectory induction often summarizes selected trajectories into holistic experience entries. Without systematically analyzing how environmental outcomes support or contradict the different factors involved in a decision, the experience induction can be unreliable and inefficient.

% 区别于别的从Trajectories induce experience 并做相关性检索获取经验的方法，TOE的范式下，当一个结果发生之后，outcome可以迅速的向所有相关experience进行反向传播，因为经验本身就与reasoning同结构，可以做到完美的Attributability。相比较之下，传统的inter-trajectory induction方法与环境交互的结果只能部分作用于相关的经验，无法做到经验系统性的更新。 In terms of \emph{Attributability}, current abstraction mechanisms summarize only selected or semantically related trajectories, without systematically analyzing how environmental outcomes support or contradict the different factors involved in a decision. This can lead to incomplete or unreliable experience induction.

\textbf{Transferability.} The progressive decomposition of reasoning perspectives improves experience transfer across heterogeneous tasks. Rather than matching a new problem against an entire historical instance, ToE retrieves experience at the level of individual perspectives or partial reasoning paths. This distinction is important in low-repetition settings, where task instances may be semantically dissimilar but require similar analytical logic. For example, financial news varies substantially across days, while the reasoning perspectives used to assess sentiment, fundamentals, or market conditions may remain transferable. ToE reuses \emph{how a problem should be analyzed}, rather than relying on surface-level similarity between problems.

% This progressive decomposition of reasoning perspectives substantially improves the transferability and reusability of experience. When encountering a new but related problem, the agent no longer needs to rely on coarse-grained matching against an entire problem instance as the original problem could be 语义上很复杂，我们要复用的经验是分析问题的角度，而不是问题本身，因为在一些场景，任务的重复性很低，如在金融新闻情绪分析任务中，每天的新闻千差万别，我们要复用的经验不是相似的新闻，而是相似的分析逻辑。
%  Instead, it can retrieve prior experience at the level of local reasoning perspectives or partial tree paths. This enables more precise reuse of effective reasoning patterns while allowing the experience units that perform well on the current task to be continually reinforced and refined.
\textbf{Evolvability.} 
%简述经验的生成、合并、删除等优势。
ToE organizes experience into manageable nodes and paths, supporting the addition of new analytical perspectives, the reliability calibration of existing paths, and the consolidation of redundant nodes.
\textbf{Efficiency.} 
 Compared with methods that repeatedly induce experience from related trajectories until inference time, ToE organizes experience before inference in a hierarchical structure. This narrows retrieval to relevant branches and limits updates to affected nodes, reducing storage, computational, and contextual overhead.  The resulting experience can therefore be generated, maintained, retrieved, and composed more efficiently.
%经验组织上的优越性使得更新和retrieve都变得很高效
% Compared with methods that induce experience from related trajectories until inference time, our method 事先将经验进行组织化，其hierarchical的组织结构使得经验的生成、更改和调用都天然的更加高效。

% Thus，our contributions are summarized as follows:\
% \indent (1) We propose \textbf{ToE}, a structured experience-management framework that organizes experience by hierarchical reasoning perspectives.\

% \indent (2) 验证了在24点这种complex problem solving 问题上的优越性
% \indent (3) 在\textbf{FinEvolveBench}, 这种 low-repetition tasks with delayed and implicit feedback数据集上，验证了我们方法的优越性。
% \begin{figure*}[t]
% \centering
% \includegraphics[width=\textwidth]{figures/Fig0_diff_exp_compare_v2.pdf}
% \caption{
% Comparison of experience-management paradigms.
% (a) Intra-trajectory transformation stores or transforms individual task trajectories;
% (b) inter-trajectory induction distills reusable knowledge from multiple past trajectories;
% and (c) Tree-of-Experience (ToE) organizes experience as hierarchical analytical perspectives and retrieves a structured reasoning path for inference.
% }
% \label{fig:diffTask}
% \end{figure*}
Thus, our contributions are summarized as follows:
\begin{itemize}
\item We propose \textbf{T}ree-\textbf{o}f-\textbf{E}xperience (ToE), a structured experience-management framework that aligns experience organization with the hierarchical reasoning process of LLM agents, thereby improving attributability, transferability, evolvability, and efficiency.

\item We reveal a critical limitation of conventional experience-management methods in complex problem-solving tasks: their inability to accurately attribute outcome feedback can reinforce unreliable experience and even degrade performance relative to experience-free baselines.

\item We validate ToE on two complementary benchmarks. On \textsc{Game of 24}, ToE achieves 85.3\% accuracy, outperforming the experience-free ToT baseline by 20.4 percentage points while reducing LLM calls by 78.7\%. On \textsc{FinEvolveBench}, a low-repetition benchmark with delayed, implicit, outcome-level feedback, ToE improves \textbf{tsIC} by an average of 41.24\% over the experience-free pipeline across 12 evaluation settings.

\end{itemize}

\section{Related Work}
Prior work enables LLM agents to reuse interaction history through episodic storage or experience abstraction. Episodic methods preserve complete or partial trajectories; for example, Synapse \citep{zheng2024synapse} retrieves abstracted state--action trajectories as in-context exemplars. Although such representations retain rich context, they also entangle transferable knowledge with instance-specific details and make outcome-level feedback difficult to attribute to individual reasoning components.

To improve transferability, intra-trajectory methods transform individual interactions into compact experience. Reflexion \citep{shinn2023reflexion} converts execution feedback into verbal reflections, while MemRL \citep{memrl2026} summarizes each trajectory into episodic memory and continually calibrates its utility through environmental feedback for value-aware retrieval.However, independently derived memories often remain fragmented and context-dependent. Inter-trajectory methods instead induce shared patterns across multiple trajectories. ReMe \citep{cao2026remember} extracts fine-grained procedural knowledge through success-pattern recognition, failure analysis, and comparative distillation, and SkillRL \citep{xia2026skillrl} recursively distills and organizes reusable skills. While these approaches improve generalization, the induced experience is typically derived from selected trajectory groups without a unified structure, making accumulated experience fragmented and incomplete.

Several methods explicitly structure agent memory. MetaFlowLLM \citep{fan2026generalizing} organizes reusable executable workflows into a hierarchical tree, Mem0 \citep{chhikara2025mem0} models entities and relations through a memory graph, and G-Memory \citep{zhang2026g} structures multi-agent experience at multiple abstraction levels. Their organizational semantics, however, center on workflows, entity relations, or multi-agent interactions. In contrast, ToE structures experience as analytical perspectives aligned with LLM reasoning and calibrates their reliability through environmental feedback. This supports feedback attribution, systematic experience updating, perspective-level transfer, and efficient retrieval, particularly under low task repetition and ambiguous outcome attribution.
\section{Methodology}
\label{sec:methodology}
Inspired by Tree of Thoughts (ToT), which enables language models to explore multiple reasoning paths over intermediate thoughts \citep{yao2023tree}, we construct the experience repository by mirroring the reasoning process of LLM agents. As shown in Fig.~\ref{fig:diffTask}, we organize the experience pool $\mathcal{E}$ as a depth-constrained yet width-expandable tree, which can be denoted as:
% We organize the external experience pool $\mathcal{E}$ as a depth-constrained yet width-expandable tree, in which each root-to-leaf path encodes an executable analytical perspective. An illustrative example is provided in Fig.~\ref{fig:toe_case}.
\begin{equation}
\mathcal{E} =
\left\{ \Big( \mathbf{\Pi}_i, \mathbf{Q}_i, \mathbf{M}_i \Big) \right\}_{i=1}^{|\mathcal{E}|}.
\end{equation}

\begin{figure}[t]
    \centering
    \includegraphics[width=\columnwidth]{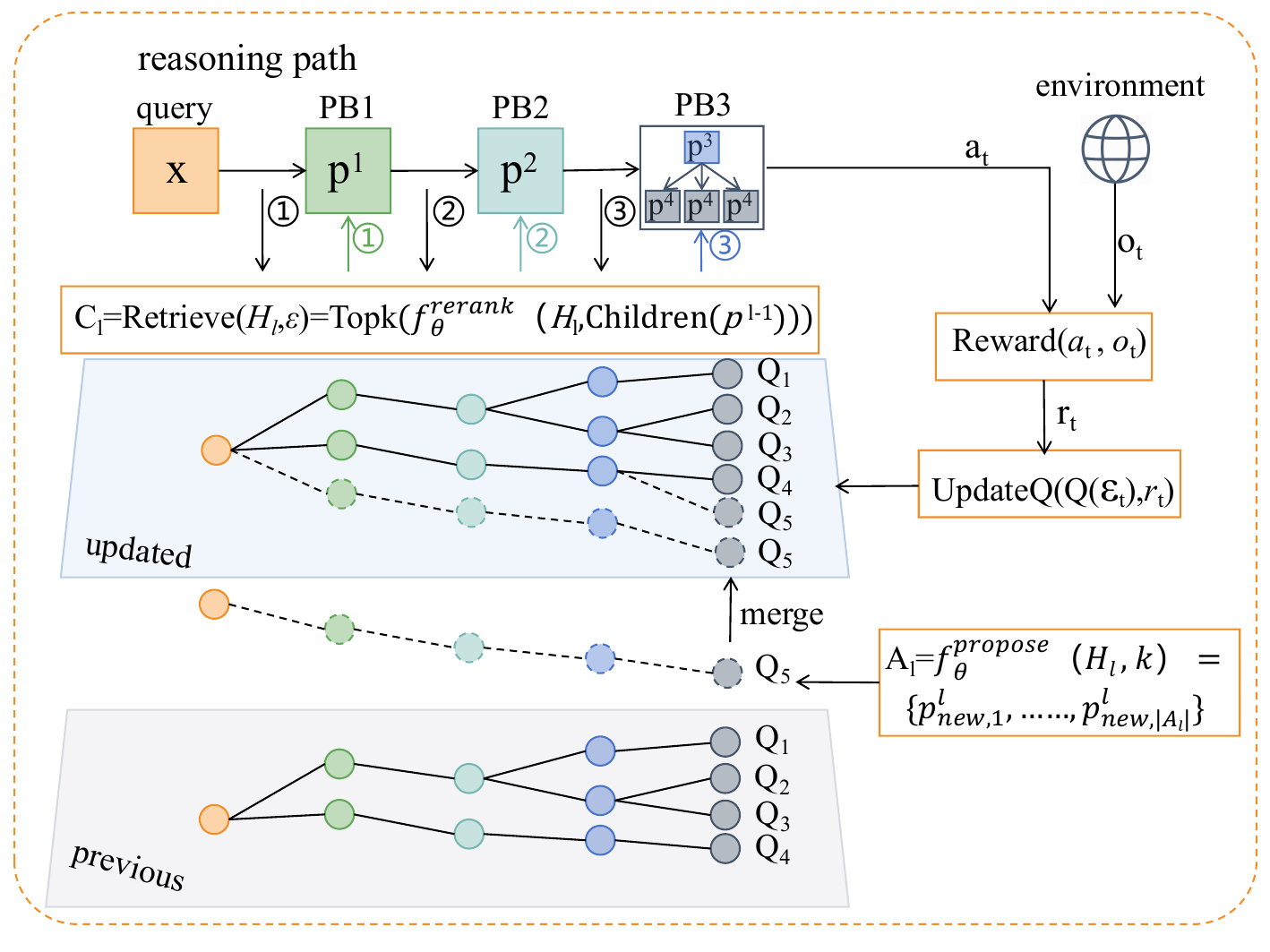}
    \caption{ Overall framework of ToE. Illustration of the reasoning and experience-evolution process, including hierarchical experience retrieval, environment-based feedback, reliability updating, and experience expansion and merging.
    }
    \label{fig:toe_experience_evolution}
\end{figure}

Each tuple $e_i = (\mathbf{\Pi}_i, \mathbf{Q}_i, \mathbf{M}_i) $ denotes a specific experience that consists of a depth-$L$ analytical path $\mathbf{\Pi}_i = (p_i^{(1)}, \dots, p_i^{(L)})$, a feedback-derived reliability state $\mathbf{Q}_i$, and maintenance metadata $\mathbf{M}_i$.
% For example, As shown in Fig.~\ref{fig:diffTask}, XXX High-level factors, such as ``policy planning'', are used as root nodes, while more specific analytical directions, such as ``supply-side administrative constraints'', are organized as lower-level nodes.

% As shown in Fig.~\ref{fig:diffTask}, ToE hierarchically organizes analytical perspectives from higher-level analytical perspectives to lower-level analytical perspectives. For example, \textit{government policy} is refined into \textit{cash support} and \textit{higher spending}. This is only for concise, the stored experience might be government policy has a new policy cash support to 促进消费，this positive sentiment has a reliabilty of 1.15 which is somewhat 确性。
% Another analytical perspective start from \textit{company fundamentals} is refined into \textit{profitability} and \textit{revenue growth}. This experience might be from company fundamental, profitability 能力下降， revenue growth可能减少，this negative sentiment has a reliabilty of 0.2, 因为降价并不一定导致盈利能力下降，而环境反馈会纠正这条错误的经验。
As shown in Fig.~\ref{fig:diffTask}, ToE organizes analytical perspectives hierarchically from general categories to more specific sub-perspectives. Given the news that the government provides shopping subsidies while major retailers reduce product prices, the agent may analyze retail sentiment from two perspectives: \textit{government policy} and \textit{company fundamentals}. The former can be refined into \textit{cash support} and \textit{higher spending}, yielding an experience that government subsidies stimulate consumption and imply positive sentiment, with a reliability score of 1.15. The latter can be refined into \textit{profitability} and \textit{revenue growth}, yielding a negative assessment that price reductions may weaken profitability and revenue growth. However, through continued interaction with the environment, this experience may be assigned a low reliability score of 0.2, as price reductions do not necessarily lead to lower profitability.

We posit that aligning experience organization with the reasoning structure of LLM agents improves outcome attribution, cross-task transfer, systematic experience updating, and retrieval efficiency. Realizing these benefits requires ToE to address three key questions. First, \textbf{reasoning granularity}: how should reasoning perspectives be defined to remain sufficiently abstract for cross-task transfer while providing concrete guidance for problem solving? Second, \textbf{reasoning-path selection and expansion}: given a new task, how should the agent select an appropriate path from the existing experience tree or extend the tree with new reasoning perspectives? Third, \textbf{experience-tree evolution}: how should experience be accumulated, reliability-calibrated through environmental feedback, and selectively merged or pruned while keeping the tree compact, effective, and maintainable? 
We address these three questions in the following subsections.

% The overall framework of TOE is illustrated in Fig. \ref{}. Depending on how we decompose the problem into subproblems, TOE 会跟随reasoning path recursively 去 experience tree里面寻找适合当前推理节点的experience（retrieval 模块），用于进一步推理。如果没有合适的已有经验，我们会有proposer模块propose新的经验路径，在这同时有merge模块maintain exp tree 的 efficiency。当经验被使用，在环境中验证之后，会有updateQ 模块将outcome系统性地归因给曾取用的经验。
%  \begin{figure}[!t]
%     \centering
%     \includegraphics[
%         width=\columnwidth,
%         trim=0 0 0 8mm,
%         clip
%     ]{figures/toe-radar.pdf}
%     \caption{\textbf{tsIC} results on \textsc{FinEvolveBench} across backbone models, exploitation stages, and prediction horizons.}
%     \label{fig:main_results}
% \end{figure}

% \begin{figure}[t]
%     \centering
%     \includegraphics[width=\columnwidth]{figures/new.pdf}
%     \caption{ Overall framework of ToE. Illustration of the reasoning and experience-evolution process, including hierarchical experience retrieval, environment-based feedback, reliability updating, and experience expansion and merging.
%     }
%     \label{fig:toe_experience_evolution}
% \end{figure}

%我们会在接下来的行文中具体解释回答这三个问题
The overall framework of ToE is illustrated in Fig.~\ref{fig:toe_experience_evolution}. Given a problem decomposition structure, ToE follows the agent's reasoning path and recursively retrieves experiences relevant to the current reasoning node from the experience tree through the retrieval module. When no suitable experience is available, the proposer generates a new experience path, while the merger consolidates semantically overlapping nodes to maintain a compact and efficient tree. After the retrieved experience is applied and evaluated through environmental interaction, the UpdateQ module systematically attributes the observed outcome to the experiences involved and calibrates their reliability accordingly.

% $\mathbf{P}_i = (p_i^{(1)}, \dots, p_i^{(L)})$ is a depth-$L$ experience path, whose upper levels encode abstract task patterns and lower levels encode concrete reasoning principles.
% %New experiences follow the same interface by adding leaf nodes.
% $\mathbf{Q}_i$ stores experience-specific utility estimates for the path. In our settings, it illustrates how important or correct is this experience's analytical perspective.
% $\mathbf{M}_i$ contains non-semantic metadata, such as recall/hit counts and identifiers, used only for bookkeeping and utility updates.
% We provide an example experience tree in the Appendix to illustrate the hierarchical organization of experiences. High-level factors, such as ``policy planning'', are used as root nodes, while more specific analytical directions, such as ``supply-side administrative constraints'', are organized as lower-level nodes.
% 

% --- 20270720
% \subsubsection{Granularity of Analytical Perspectives}
% \subsubsection{Constructing Analytical-Perspective Nodes}

% Trajectory-based memory methods typically store and reuse the complete reasoning trace generated by a language model as an indivisible unit. In contrast, ToE explicitly decomposes the reasoning process into a set of structured and reusable \emph{analytical perspectives}, according to the structure of the task and its reasoning requirements. Experience storage, retrieval, and updating are subsequently performed around these perspectives rather than entire reasoning trajectories. 

\subsubsection{Granularity of Analytical Perspectives}
The structure of the experience tree is not a predefined template. Instead, its semantics and granularity are jointly determined by the problem structure, the task objective, and the information required during problem solving. For example, in mathematical reasoning tasks such as the Game of 24, analytical perspectives may correspond to \textit{operator selection} or \textit{number grouping}. In financial investment tasks, they may instead capture dimensions such as the \textit{macroeconomic environment}, \textit{company fundamentals}, or \textit{geopolitical factors}.

Within ToE, analytical perspectives at different levels form a progressive hierarchy from abstract reasoning dimensions to concrete decision criteria. A higher-level node describes a relatively broad analytical dimension, whereas each lower-level node inherits the semantics of its parent and further specializes a particular subproblem, causal mechanism, or reasoning basis. A root-to-leaf path represents a progressive reasoning process that begins with a general analytical direction and gradually specializes it into concrete evidence or decision criteria.

In ToE, the organization of experience is aligned with the reasoning structure of the LLM agent. Accordingly, the semantics and granularity of the experience hierarchy are jointly determined by the problem structure, task objective, and information required during problem solving. The initial hierarchy can be constructed in two ways. For tasks with a well-defined reasoning procedure, it can be manually specified based on prior knowledge of the task. Alternatively, it can be induced from multiple model-generated reasoning paths by recursively clustering and summarizing similar perspectives at successive levels.

We distinguish between two forms of problem structure. In a \textit{sequentially dependent} structure, solving each subproblem depends on the result of the preceding one, and experience is retrieved separately at each reasoning step. For example, following the ToT \citep{yao2023tree} formulation of \textsc{Game of 24}, the reasoning process can be divided into three stages corresponding to states with four, three, and two remaining numbers, respectively. ToE retrieves an appropriate experience at each stage. By contrast, in an \textit{independent parallel} structure, the solution requires multiple complementary perspectives that can be considered jointly. In \textsc{FinEvolveBench}, ToE retrieves multiple relevant reasoning paths from the experience tree simultaneously and integrates them within a single inference. For such tasks, the initial hierarchy can be induced by progressively clustering and abstracting collected reasoning paths, and is subsequently refined during continual evolution.

The construction of a new node should satisfy three basic rules. First, a child node must maintain a clear semantic subsumption relation with its parent: its content should extend, specialize, or operationalize the analytical perspective represented by the parent node. Second, the new node must introduce information with genuine discriminative value, enabling the model to derive additional conclusions, distinguish among candidate solutions, or revise an existing judgment. Third, the granularity of a node should remain balanced. An overly broad node provides insufficient guidance for subsequent reasoning, whereas an excessively fine-grained node reduces the likelihood that the corresponding experience can be reused across tasks.

Accordingly, a new child node is introduced when the current analytical perspective still contains multiple relatively independent subproblems and further decomposition is expected to improve reasoning efficiency or cross-task experience reuse. Expansion terminates when additional decomposition no longer contributes new reasoning-relevant information, or when the current node is sufficiently specific to directly support a local inference or decision.

\subsubsection{Reasoning Path Selection and Generation}
 % +XXX 大模型推理相关文献 autoregressive 的LLM要具有结构化思维的能力需要XXXX，COT、TOT等文献。而TOE给了这种思维能力以进化的能力。Solving a complex problem requires a coherent and progressively refined reasoning process. In our TOE framework, the retrieval of experience 与大模型的推理过程同步进行。
Autoregressive LLMs rely on structured intermediate reasoning to solve complex problems coherently and progressively. For example, Chain-of-Thought (CoT) organizes such reasoning into sequential steps, while Tree of Thoughts (ToT) extends it by exploring multiple candidate reasoning paths \citep{wei2022chain,yao2023tree}.
Building on these structured reasoning paradigms, ToE enables the reasoning process to evolve through accumulated environmental experience. Experience retrieval proceeds synchronously with LLM reasoning, allowing the model to draw not only on previously explored analytical perspectives but also on their feedback-derived reliability.
% Depends on different problems, a reasoning progress could be 逐步深入、一步步解决，如同24点问题，我们先考虑选取2个数进行加减乘除操作，再考虑得到的结果与剩下2个数是否能凑成24点。也可以是多方面的思考，如同在金融情绪分析场景，我们需要并行的对多个角度进行思考，对多个角度进行发散性思考。
As shown in Fig. \ref{fig:toe_experience_evolution}, experience retrieval in ToE proceeds synchronously with the LLM's reasoning process. Depending on the task, reasoning may proceed through sequentially dependent subproblems, as in the Game of 24, or across multiple independent perspectives, as in financial analysis. In the former case, ToE repeatedly traverses the experience tree as reasoning advances step by step; in the latter, it retrieves multiple relevant reasoning paths to support complementary analysis. In both cases, the retrieved experience guides subsequent reasoning and identifies promising directions for further exploration.

% 具体的，Given the original task input $x$, 假设我们当下的思考步骤为l，first retrieves $l=1$ level experience nodes relevant to the current problem using the retrieval function $\operatorname{Retrieve}(\cdot)$:
% %
% \begin{equation}
%    \mathcal{C}_l = \operatorname{Retrieve}(\{x，\Pi^{l-1}\}  , \mathcal{E}),
% \end{equation}
% %
% where $\mathcal{E}$ the experience tree and $\mathcal{C}_l$ denotes the retrieved candidate set. $\Pi^{l-1}$ denotes the realized reasoning path of the agent.

Specifically, given a task input $x_t$ and the reasoning path constructed before step $l$, we define the current reasoning context as
\begin{equation}
H_l=(x_t,\Pi^{l-1}),
\end{equation}
where $\Pi^{l-1}=(p^1,\ldots,p^{l-1})$ denotes the reasoning path established before step $l$. ToE retrieves candidate nodes from the experience tree $\mathcal{E}$ by reranking child nodes of $p^{l-1}$ and selects the top-k ranked perspectives:
\begin{equation}
\begin{aligned}
\mathcal{C}_l
&= \operatorname{Retrieve}\left(H_l,\mathcal{E}\right) \\
&= \operatorname{TopK}\left(
f_{\theta}^{\mathrm{rerank}}
\left(
H_l,\operatorname{Children}(p^{l-1})
\right)
\right).
\end{aligned}
\end{equation}
The prompt-based LLM reranker $f_{\theta}^{\mathrm{rerank}}(\cdot)$ assesses each candidate for task relevance and consistency with the existing reasoning path, filtering out irrelevant, redundant, or incompatible perspectives in the meantime. 
If no retrieved candidate provides a valid continuation, ToE invokes a prompt-based LLM proposer
that generates at most $k$ new analytical perspectives.
\begin{equation}
\mathcal{A}_l = f_{\theta}^{\mathrm{propose}}
\left(H_l,k\right)
=
\{p_{\mathrm{new},1}^l,\ldots,p_{\mathrm{new},|\mathcal{A}_l|}^l\}.
\end{equation}

Each proposed perspective extends the current analytical path as
\begin{equation}
\Pi_{\mathrm{new},j}^{l} = 
\Pi^{l-1}\mathbin{|}p_{\mathrm{new},j}^{l},
\end{equation}
where $\mathbin{|}$ denotes path concatenation. 
The agent will then traverse all the possible $\Pi_{\mathrm{new},j}^{l}$ at step $l$.
When $l=L$ or the stopping criterion is satisfied, ToE assigns a neutral reliability prior, $\mathbf{Q}_i=q_0$, to a newly constructed analytical path. For a previously observed path, ToE retrieves its stored reliability state $\mathbf{Q}_i$ from the experience pool to inform the current solution.

% \paragraph{Utility-Conditioned Experience Utilization.}

% After an experience path is selected, the original query is augmented with the selected experience and its utility. The augmented query is defined as

% \begin{equation}
% \widetilde{x}=Augment(x,P^{*},Q(P^{*})).
% \end{equation}
% The function in the equation denotes the experience augmentation process. It combines the original query, the selected experience path, and the corresponding utility. The model output is then generated by
% \begin{equation}
% \hat{y}=f_{\theta}(\widetilde{x}).
% \end{equation}

% The utility is used only after the experience path has been selected. It does not participate in experience recall, candidate reranking, or reasoning-path selection. ToE does not prescribe how the selected experience and its utility are combined. Their concrete usage is determined by the task-specific implementation of the augmentation function.
We denote $\mathcal{E}_t= \{\Pi_L,Q(\Pi_L)\}$ as
the retrieved experience set for $x_t$. the final answer is generated by conditioning the LLM on the task input, the selected analytical paths, and its reliability state:
\begin{equation}
a_t =f_{\theta}(x_t,\mathcal{E}_t).
\end{equation}

\subsubsection{Experience Tree Evolution}
% %最后，我们解决  how should experience
% be accumulated, consolidated, updated, and pruned through continual environmental interaction while keeping the tree compact, effective, and maintainable这个问题。
Finally, we address how experience should be accumulated, reliability-calibrated through outcome feedback, and selectively merged or pruned through continual interaction with the environment to keep the tree compact, effective, and maintainable.

Following the experience utilization process defined above, the model generates the answer $a_t$ based on the query augmented with the selected experience path and its reliability. Once the corresponding environmental feedback becomes available, ToE uses this feedback to evaluate the selected path and update the experience tree.

After the task is completed, the system receives delayed feedback from the environment. The feedback is converted into an experience-level update signal:
\begin{equation}
r_{t} = \psi_{\tau}(a_t,o_{t+h}),
\end{equation}
where $a_t$ denotes the model output or action generated using the selected experience, $o_{t+h}$ denotes the delayed task outcome, and $\psi_{\tau}$ denotes the task-specific feedback function.
%For example，在金融场景我们用的 reward 是什么，这里要交代一下，因为你 experiment 里面可能没有写吧？
% For example, in financial tasks, $r_{\mathrm{env}}$ is instantiated as the market-adjusted return, computed as the sector's closing price minus its opening price, divided by the opening price, and further adjusted by subtracting the corresponding return of a broad-market index.
For example, in the FinEvolveBench task, we use $r_{\mathrm{env}} = a_to_{t+h}$ since predictive performance is evaluated by the information coefficient (IC).

% 当待更新的思考路径中，存在LLM新purpose的节点时，会通过embedding召回经验树中相似的思考节点，并采用Prompt-Based LLM方法，让模型判断新增的经验是否应与现有经验合并。这里是一方面是为了避免在purpose阶段模型错误的提出了相似的分析角度，另外一方面是为了防止在待更新周期内，其他的轨迹提出了相似的分析角度更新后，产生了冗余节点。再一方面是为了防止同一个父节点中存在过多相似节点。

% When the selected experience path contains a new analytical-perspective node, ToE retrieves similar nodes via embedding-based search and uses a prompt-based LLM to determine whether to merge them, preventing redundant perspectives across trajectories. If matched, their experiences and feedback statistics are consolidated; otherwise, the new node is inserted at the corresponding position.

% When the analytical-perspective path used for the current task contains a new node, ToE retrieves semantically similar nodes via embedding search and uses a prompt-based LLM to determine whether the new node should be merged with an existing one. This step removes near-duplicate perspectives introduced during path construction or concurrent updates and limits redundancy among sibling nodes. If matched, the new node and its associated experience statistics are consolidated into the existing node; otherwise, it is inserted under the corresponding parent.

For all used experiences associated with this outcome, experience reliability is then updated according to an update mechanism.
\begin{equation}
Q(e) \leftarrow \operatorname{UpdateQ}\left(Q(e),r_{t} \right),  \qquad
\forall e\in\mathcal{E}_t.
\end{equation}
%对于不同的task，需要设计不同适合的update mechanism，这里我们例举两个可行的方案，We consider two implementations of $\operatorname{UpdateQ}(\cdot)$: a formula-based update and an LLM-based update.
The appropriate reliability update mechanism may vary across tasks. We therefore present two representative implementations of $\operatorname{UpdateQ}(\cdot)$: a formula-based update and an LLM-based update.

\paragraph{Formula-based update.}
% Each experience $e$ maintains a utility vector
% $\mathbf{Q}(e) \in [0,2q_0]^d$, where $q_0$ is the neutral prior. This value 反映了经验的可靠性，在经验的使用时 controls the activation strength of the experience. 更新的步伐与环境反馈成正比。当outcome给了正向反馈时， $\mathbf{Q}(e)$应该增大，负向反馈时验证了经验时，这个值应该减小。在经验积累的前期反馈较少时，更新的步伐应较大，而在反馈逐渐增多时，一次的影响应该衰减。当$Q$偏离$q_0$较远时，系统对经验的确定性较高，更新的脚步变小，而$Q$离$q_0$较近，对这个经验的可用性不太确定时，一次更新的步伐变大。

Each experience $e$ maintains a reliability value $\mathbf{Q}(\mathcal{E}_t)\in[0,2q_0]^{|\mathcal{E}_t|}$, where $q_0$ denotes the neutral prior. The reliability value reflects the empirical validity of the experience and controls its activation strength during use. Positive environmental feedback increases the corresponding reliability dimensions, whereas negative feedback decreases them.

Given delayed environmental feedback $\mathbf{r}_{\mathrm{env}}$, the reliability of each attributed experience is updated as
\begin{multline}
\mathbf{Q}(\mathcal{E}_t) \leftarrow \mathbf{Q}(\mathcal{E}_t)
+ \eta
\frac{
1-\left(\frac{\mathbf{Q}(\mathcal{E}_t)-q_0}{q_0}\right)^2
}{
1+\log(1+n_{\text{hit}})
}
\cdot r_{t}, 
%  \qquad
% \forall e\in\mathcal{E}_t,
\end{multline}
where $\eta$ is a learning rate parameter, $n_{\mathrm{hit}}$ is the historical usage count of experience $e$, and $\mathcal{E}_t$ denotes the experiences attributed to the current outcome.
The update magnitude is adaptive and proportional to the feedback strength. The usage-dependent denominator reduces the influence of individual feedback as evidence accumulates, while the boundary-aware factor produces larger updates near the neutral prior and smaller updates near the reliability bounds. Consequently, repeatedly ineffective experiences are gradually assigned lower activation strengths rather than being explicitly removed, yielding a form of soft forgetting.

\paragraph{LLM-based update.}
Alternatively, an LLM-based updater can analyze the observed outcome and assign experience-specific update magnitudes according to each experience's estimated contribution. This enables more fine-grained credit assignment, but may also introduce subjective biases from the LLM.
\begin{equation}
\mathbf{Q}(\mathcal{E}_t) \leftarrow
f_{\theta}^{\mathrm{update}}\bigl(\mathcal{E}_t, \mathbf{Q}(\mathcal{E}_t), \mathbf{r}_{\text{env}}\bigr).
\end{equation}
Here, $f_{\theta}^{\mathrm{update}}(\cdot)$ denotes a prompt-based LLM updater whose behavior is constrained by predefined update rules encoded in the prompt.

\paragraph{Maintenance.}
Here, we describe how ToE keeps the experience tree compact and effective during continual updates. First, the generation of new nodes is tightly controlled in both quantity and quality, limiting unnecessary expansion at its source. Second, as part of its routine update procedure, ToE retrieves semantically similar existing nodes through embedding-based search and employs a prompt-based LLM merger to determine whether some node should be merged. This strategy reduces redundancy arising from repeated perspective construction, concurrent updates, and semantic overlap among sibling nodes. We do not prune experiences solely because their reliability falls below a predefined threshold, as low-reliability experiences may still provide useful negative evidence by identifying unreliable reasoning patterns and preventing repeated errors. In addition, each experience maintains metadata $\mathbf{M}_i$, including its recall count, hit count, node index, and creation timestamp. These statistics provide a foundation for more advanced metadata-aware experience maintenance, which we leave for future work.

% 数据结果表格不在正文中
\section{Experiments}
% XXXXXXXWe evaluate our method on two datasets: 24Game (a XXXXXdataset...) and Fine-Evolved Bench (a XXXXdataset...)

We evaluate our method on two complementary tasks: \textsc{Game of 24}\citep{yao2023tree}, and the \textsc{FinEvolveBench} \citep{deng2026finevolve}. \textsc{Game of 24} is a mathematical reasoning challenge that requires combining four given numbers using basic arithmetic operations ($+,-,\times,/$) to obtain 24; for example, given 4 9 10 13, a valid solution is $(10-4)\times(13-9)=24$. FinEvolveBench is a sentiment analysis benchmark, which is required to analyze temporally ordered financial news to predict future market sentiment for Chinese A-share industry indices.

\begin{table*}[!b]
    \centering
    \begingroup
    \setlength{\tabcolsep}{2.8pt}
    \renewcommand{\arraystretch}{1.18}

    \resizebox{\textwidth}{!}{%
    \begin{tabular}{@{}l*{12}{c}@{}}
        \toprule

        & \multicolumn{6}{c}{DeepSeek-V4-Flash}
        & \multicolumn{6}{c}{Qwen3.6-35B-A3B} \\

        \cmidrule(lr){2-7}
        \cmidrule(lr){8-13}

        \textbf{Method}
        & \multicolumn{2}{c}{10d}
        & \multicolumn{2}{c}{20d}
        & \multicolumn{2}{c}{40d}
        & \multicolumn{2}{c}{10d}
        & \multicolumn{2}{c}{20d}
        & \multicolumn{2}{c}{40d} \\

        \cmidrule(lr){2-3}
        \cmidrule(lr){4-5}
        \cmidrule(lr){6-7}
        \cmidrule(lr){8-9}
        \cmidrule(lr){10-11}
        \cmidrule(lr){12-13}

        &
        Exploitation & Overall
        & Exploitation & Overall
        & Exploitation & Overall
        & Exploitation & Overall
        & Exploitation & Overall
        & Exploitation & Overall \\

        \midrule

        Pipe
        & 0.0487 & 0.0243
        & 0.0871 & 0.0517
        & 0.0843 & 0.0595
        & 0.0458 & \textbf{0.0349}
        & 0.0746 & 0.0676
        & 0.0851 & 0.0677 \\

        Pipe+mem0
        & 0.0393 & 0.0202
        & 0.0608 & 0.0325
        & 0.0762 & 0.0288
        & 0.0379 & 0.0132
        & 0.0567 & 0.0336
        & 0.0747 & 0.0475 \\

        Pipe+MemRL
        & 0.0473 & 0.0175
        & 0.0449 & 0.0182
        & 0.0671 & 0.0269
        & 0.0159 & 0.0186
        & 0.0356 & 0.0481
        & 0.0266 & 0.0135 \\

        Pipe+ToE
        & \textbf{0.0707} & \textbf{0.0313}
        & \textbf{0.1311} & \textbf{0.0741}
        & \textbf{0.1456} & \textbf{0.0899}
        & \textbf{0.0753} & 0.0322
        & \textbf{0.1387} & \textbf{0.0749}
        & \textbf{0.1180} & \textbf{0.0753} \\

        \bottomrule
    \end{tabular}%
    }

    \endgroup
    \caption{\textbf{tsIC} results over 10-, 20-, and 40-trading-day prediction horizons during the Exploitation and Overall periods of \textsc{FinEvolveBench}, using DeepSeek-V4-Flash and Qwen3.6-35B-A3B as the backbone LLMs.}
    \label{tab:horizon_ablation}
\end{table*}

\label{sec:experiments}

\subsection{Experimental Setup}
We briefly introduce our experimental setup and include implementation details and more experimental results in the Appendix due to space limitations. Code is also included in the supplementary material.
Our main experiments use \textsc{DeepSeek-V4-Flash} as the backbone model, with \textsc{Qwen3.6-35B-A3B} included as an additional backbone-model ablation. 
%The context window is set to 32,768 tokens, and we use the officially recommended decoding configuration. To reduce the effect of randomness from a single run, each method is run independently three times, and the tables report the average performance across the three runs.
% Across all experiments, we set the experience-tree depth $L=3$, the first-level candidate-retention size to $k_1=8$, the leaf-level retrieval cosine-similarity threshold to $\lambda=0.8$, the neutral utility to $q_0=1$, and the learning rate to $\eta=10$. 

% -----------

\paragraph{Compared methods.}
% We compare our method, with three SOTA(这里我不能确定所谓的 SOTA 是什么，只能说这些工作非常的新，都是近几年的最新出炉的 )work experience works,mem0}\cite{chhikara2025mem0}, 
% MemRL}\citep{memrl2026}, 和ReMe}\citep{cao2026remember}

% 在 24 Game这个数据集上, all methods use Tree-of-Thoughts (ToT) as the underlying reasoning procedure. We use experience-free \textbf{ToT} as baseline, and compare our method with 另外3个方法。
% On \textsc{FinEvolveBench} 数据集上,我们用这个数据集提供的pipeline 方法\textbf{Pipe} 作为baseline,将我们的方法与mem0和MemRL方法对比。 We exclude ReMe from this setting because, in our implementation, it constructs substantially longer experience contexts, making its inference cost impractical when combined with long-form financial news.%XXXX具体数据支撑。
% We compare \textbf{Baseline}, an average aggregation of LLM-assigned news sentiment scores\citep{wang2018combining, mohan2019stock, lopezlira2023chatgpt, wang2024mananet}; \textbf{Pipe}, which predicts from industry-filtered news; \textbf{Pipe+mem0} and \textbf{Pipe+MemRL}, which add mem0~\citep{chhikara2025mem0} and MemRL~\citep{memrl2026} as experience modules; and \textbf{Pipe+ToE}, which uses our tree-structured experience retrieval with formula-based utility updating.

We compare ToE with three recent representative experience-based methods: Mem0 \citep{chhikara2025mem0}, MemRL \citep{memrl2026}, and ReMe \citep{cao2026remember}. 
On \textsc{Game of 24}, all methods use Tree of Thoughts (ToT) as the underlying reasoning procedure. We adopt experience-free \textbf{ToT} as the baseline and compare ToE with Mem0, MemRL, and ReMe.
On \textsc{FinEvolveBench}, we use the benchmark-provided pipeline, denoted as \textbf{Pipe}, as the experience-free baseline, and compare ToE with Mem0 and MemRL.We exclude ReMe because its original day-level global extraction jointly processes all sector-level trajectories, causing substantial context redundancy and inference overhead for long-form financial news. For example, aggregating 31 sector trajectories on September 1, 2025 produced an input of approximately 324.5 K tokens, making direct evaluation impractical even with state-of-the-art long-context models.

% We exclude ReMe because, in our implementation, it produces substantially longer experience contexts, resulting in prohibitive inference costs when combined with long-form financial news.XXXXX%XXXX具体数据支撑。

\paragraph{Evaluation metrics.}
On \textsc{Game of 24}, we report accuracy to measure task performance and the average LLM calls to evaluate experience management efficiency. On \textsc{FinEvolveBench}, we follow the benchmark's evaluation protocol and use \textbf{cross-sectional IC (csIC)} and \textbf{time-series IC (tsIC)} as the primary metrics \cite{deng2026finevolve}.

% ------- 20260726
\subsection{Performance on \textsc{Game of 24}}
\label{sec:24game}

% We first evaluate ToE on the 24 Game, a controlled reasoning task with deterministic feedback, to examine whether it can extract transferable reasoning experience while reducing inference cost. All methods use Tree-of-Thoughts (ToT) as the underlying reasoning procedure. ToT denotes the base method without an experience module, while the remaining methods combine ToT with different experience-management mechanisms. We report solution accuracy and the average number of LLM calls per problem.

% \textbf{ToT} baseline with \textbf{ToT+mem0}\cite{chhikara2025mem0}, \textbf{ToT+MemRL}\citep{memrl2026}, \textbf{ToT+ReMe}\citep{cao2026remember}, and \textbf{ToT+ToE}.

As shown in Table~\ref{tab:24game_results}, our method (ToT+ToE) achieves overall higher reasoning accuracy and efficiency compared with Mem0 \citep{chhikara2025mem0}, MemRL \citep{memrl2026}, and ReMe \citep{cao2026remember}. 
% 可以看到其他的没有对复杂问题推理优化的经验管理方法在面对 24 点这种 reasoning 的数据集上，表现都有下降，也就是说经验的使用对问题的解决反而有副作用。 而我们的方法 reasoning accuracy achieves 85.3\%, outperforming ToT by 20.4 percentage points. This suggests that ToE can attribute outcomes to relevant reasoning perspectives, filter out instance-specific details and ineffective search steps, and extract reasoning patterns that transfer across different number combinations.
Notably, experience-management methods not specifically designed for complex reasoning underperform the experience-free ToT baseline on \textsc{Game of 24}, suggesting that poorly organized experience may hinder rather than facilitate reasoning.
In contrast, ToE achieves an accuracy of 85.3\%, outperforming ToT by 20.4 percentage points. This improvement suggests that ToE effectively attributes outcomes to relevant reasoning perspectives, filters instance-specific details and ineffective search steps, and transfers reusable reasoning patterns across number combinations.

% ToE also supports local utility updates based on execution feedback, allowing useful experience paths to be reinforced and ineffective ones to be suppressed. Although the current results provide only indirect evidence for attributability and evolvability, the large accuracy gain indicates that structured experience organization is more effective than flat semantic memory in this task.

\begin{table}[!h]
    \centering
    \begin{tabular}{@{}l@{\hspace{0.1em}}c@{\hspace{0.6em}}c@{}}
        \toprule
        Method & Accuracy $\uparrow$ & Avg.\ Calls $\downarrow$ \\
        \midrule
        ToT \citep{yao2023tree}
            & 64.9\% & 51.7 \\
        ToT+mem0 \citep{chhikara2025mem0}
            & 58.4\% & 53.8 \\
        ToT+MemRL \citep{memrl2026}
            & 61.6\% & 51.2 \\
        ToT+ReMe \citep{cao2026remember}
            & 61.1\% & 53.0 \\
        \textbf{ToT+ToE (ours)}
            & \textbf{85.3\%} & \textbf{11.0} \\
        \bottomrule
    \end{tabular}
    \caption{Performance and inference efficiency on the \textsc{Game of 24}.}
    \label{tab:24game_results}
\end{table}

In terms of efficiency, ToT+ToE averages only 11.0 LLM calls per puzzle, representing a 78.7\% reduction from the 51.7 calls required by ToT. This suggests that compact experience paths guide the solution search toward promising directions while avoiding low-value branches. In contrast, ReMe, MemRL, and Mem0 require call counts comparable to the original ToT pipeline and thus yield no reduction in reasoning cost.

Moreover, the number of LLM calls does not fully capture contextual overhead. In our implementation, ReMe introduces substantially longer experience contexts while retaining nearly the same number of calls, resulting in higher token consumption. This issue becomes more severe in the financial setting, where each input already contains multiple long-form news articles. Therefore, under realistic inference-cost constraints, we do not include ReMe in the FinEvolveBench comparison.

Overall, the \textsc{Game of 24} results demonstrate that ToE jointly improves attributability, transferability, evolvability, and efficiency in a setting with a well-defined reasoning structure and deterministic feedback.

\subsection{Performance on FinEvolveBench}
\label{sec:finbench}

% 相比于有well-defined reasoning structure and deterministic feedback的24 Game问题， FinEvolveBench 这个金融情绪分析问题就更为复杂一些。首先，新闻的分析角度是发散和主观的，其次金融的feedback是有滞后的、有噪声的、且更加implicit的。
Compared with the \textsc{Game of 24}, which has well-defined reasoning structures and deterministic feedback, \textsc{FinEvolveBench} presents a more challenging setting. Financial news can be interpreted from diverse and subjective analytical perspectives, while market feedback is delayed, noisy, and implicit. The results show that our method achieves the best overall performance across the evaluated metrics. Table~\ref{tab:horizon_ablation} reports the \textbf{tsIC} results over 10-, 20-, and 40-trading-day prediction horizons using two backbone LLMs: DeepSeek-V4-Flash and Qwen3-35B-A3B.

% Figure~\ref{fig:main_results} shows the results on the 20-trading-day prediction horizon. This experiment compares three factors: basic sentiment scoring (Baseline), structured news processing without experience retrieval (Pipe), and different experience-management mechanisms added to the same prediction pipeline. The additional backbone-model ablation is also included in the same figure.

% 我们还对比了2个时间段的\textbf{tsIC}，一个是全数据集阶段，在表中是overall
% ，一个是数据集1/3数据之后开始到结束的阶段，在表中是Exploitation.
% We further 细化了 \textbf{tsIC} 结果 over two evaluation periods: Cold start phase and exploitation phase. In the initial one-third of the dataset, experience tree 还在构建阶段，utility的更新还未收敛，在接下来的时间，我们定义为
% exploitation phase，此时经验树已经初具规模，reliablity的数值更新也趋近收敛。我们关注这个阶段和the full dataset, denoted as \textit{Overall}.

 \begin{figure}[!t]
    \centering
    \includegraphics[
        width=\columnwidth,
        trim=0 0 0 8mm,
        clip
    ]{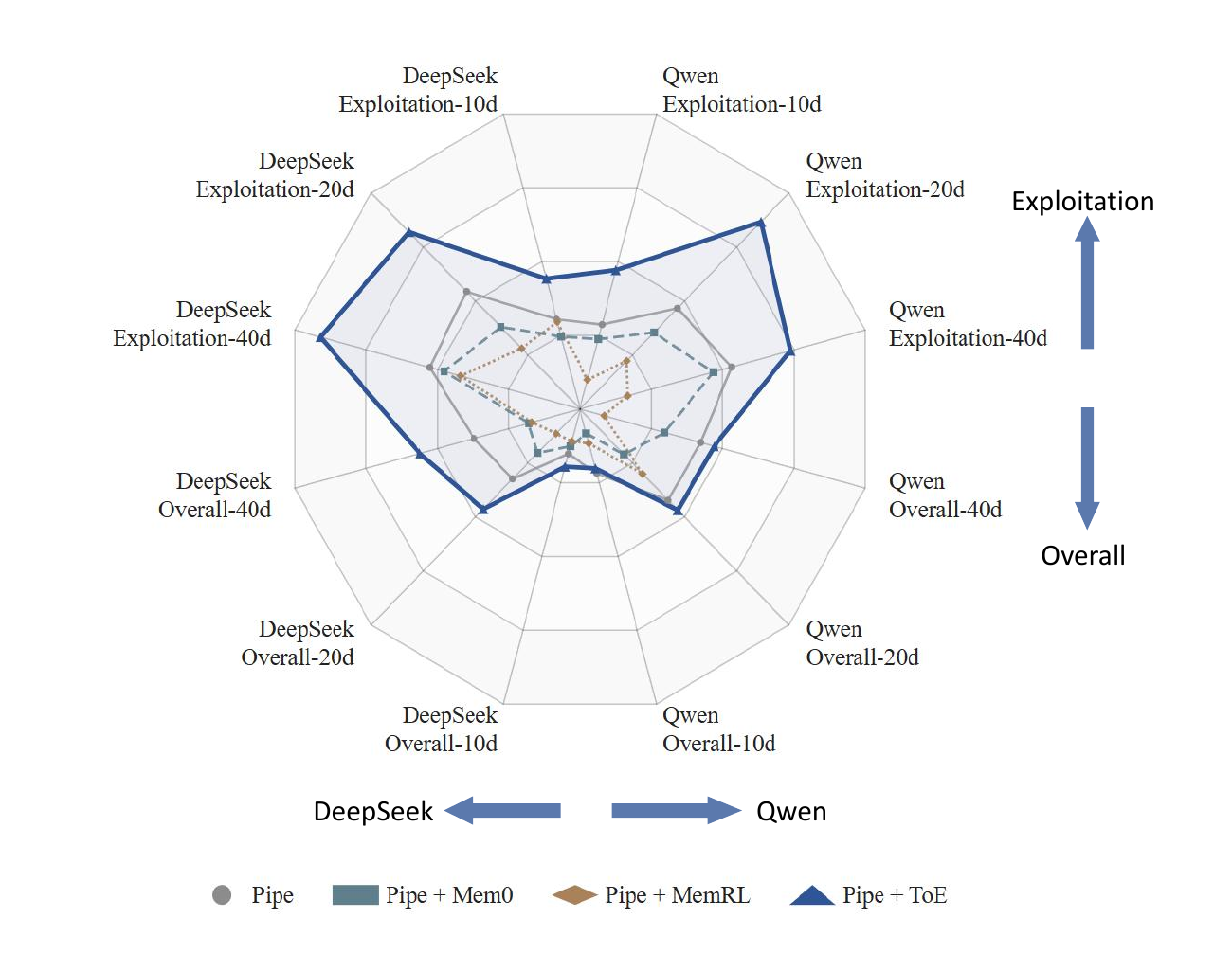}
    \caption{\textbf{tsIC} results on \textsc{FinEvolveBench} across backbone models, exploitation stages, and prediction horizons.}
    \label{fig:main_results}
\end{figure}

We further divide the evaluation timeline into two phases: a \textit{Cold-Start} phase and an \textit{Exploitation} phase. The first one-third of the dataset is designated as the Cold-Start phase, during which the experience tree is still being constructed and its reliability estimates have not yet stabilized. The remaining period is defined as the Exploitation phase, in which the experience tree has reached a meaningful scale and the reliability updates have largely converged. We therefore report the \textbf{tsIC} results for the Exploitation phase as well as for the full dataset, denoted as \textit{Overall}.
Figure~\ref{fig:main_results} presents a comparison of the key results, showing that ToE substantially outperforms Mem0 and MemRL across all settings.

% 我们可以观察到，相比于 24 点任务中这个经验只是些微地影响了 accuracy，在金融情绪分析 FinEvolve Bench 这个数据集上，当任务的 experience transferability 更低、结果更加隐晦的时候，传统的经验机制出现了失效，对结果甚至有着很大的负面影响

\begin{table}[!t]
    \centering
    \begin{tabular}{@{}lcc@{}}
        \toprule
        & \multicolumn{2}{c}{DeepSeek-V4-Flash} \\
        \cmidrule(lr){2-3}
        Strategy & tsIC & csIC \\
        \midrule
        formula-based & \textbf{0.0741} & \textbf{0.0528} \\
        LLM-based     & 0.0598 & 0.0431 \\
        \bottomrule
    \end{tabular}
    \caption{Reliability update mechanism ablation on the
    20-trading-day prediction horizon.}
    \label{tab:update_ablation}
\end{table}

Compared with the \textsc{Game of 24}, where conventional experience mechanisms only marginally affect reasoning accuracy, these methods perform substantially worse than the experience-free baseline on \textsc{FinEvolveBench}. When experience is less transferable across instances and environmental feedback is more implicit, existing memory mechanisms can even substantially degrade performance. 
% In contrast, directly incorporating Mem0 or MemRL does not consistently improve the underlying pipeline. Their average \textbf{tsIC} scores decrease to 0.0435 and 0.0317, respectively, suggesting that simply retaining or optimizing past experience is insufficient and may introduce redundant, context-dependent, or unreliable information. This observation supports the motivation of ToE: effective experience utilization requires a unified structure that organizes analytical perspectives, calibrates their reliability through outcome feedback, and selectively consolidates useful experience.

In contrast, our proposed ToE method improves the average \textbf{tsIC} by 59.57\% during the Exploitation phase and by 22.92\% over the Overall period, yielding an average relative improvement of 41.24\% across all 12 settings.
 % The improvement is particularly pronounced in the Exploitation stage, where the average \textbf{tsIC} increases from 0.0709 to 0.1132, while a consistent gain is also observed in the Overall stage, from 0.0510 to 0.0630. 

 % 对比Exploitation phase 和 overall的数据，我们可以发现experience exploitation phase 的效果明显要更好，这说明经过经验的冷启动阶段后，我们的方法要更加显著地优于其他方案。

Comparing the Exploitation-phase results with the Overall results, we observe consistently stronger performance during the Exploitation phase. This suggests that, after the initial cold-start phase in which the experience tree and reliability is  still under construction, ToE achieves a more pronounced advantage over the competing methods. 
% Moreover, the our method 在 baseline 表现更好的20- and 40-day horizons 更有优势， indicating that 在新闻预测力本身更informative 的 medium- and long-horizon prediction，我们的经验管理方法能更好的搜集信息。
Moreover, ToE delivers larger gains at the 20- and 40-day horizons, where the baseline itself exhibits stronger predictive performance. This suggests that ToE is particularly effective at identifying, accumulating, and exploiting transferable predictive patterns when financial news contains more informative signals.

The improvements are also consistent across backbone models. Relative to Pipe, ToE increases the \textbf{tsIC} by 48.61\% under DeepSeek-V4-Flash and by 33.88\% under Qwen3-35B-A3B on average, indicating that its effectiveness is not tied to a particular backbone. The only exception occurs in the 10-day Overall setting with Qwen3-35B-A3B, where ToE performs marginally worse than Pipe. Nevertheless, it still outperforms Mem0 and MemRL in this setting. The broad improvements across all remaining settings further suggest that ToE provides a more robust and transferable experience-management mechanism than existing memory-based alternatives.

Taken together, these results highlight the central advantage of ToE: it enables more effective and robust experience utilization by organizing analytical perspectives within a unified hierarchy, calibrating their reliability through outcome feedback, and selectively consolidating useful experience.

\subsection{Ablation Study on Reliability Update Mechanisms}
Table~\ref{tab:update_ablation} compares two experience-update strategies on \textsc{FinEvolveBench} under the 20-trading-day prediction horizon. Both settings use the same experience-retrieval and prediction pipeline and differ only in how experience reliability is updated. The \textit{formula-based} strategy corresponds to the Pipe+ToE configuration used in the main experiments and applies the update rule defined in Section~\ref{sec:methodology}. The \textit{LLM-based} strategy instead prompts the LLM to directly adjust the reliability score of each experience according to the observed feedback.

The formula-based strategy outperforms direct LLM-based updating on both \textbf{tsIC} and \textbf{csIC}. This suggests that, on \textsc{FinEvolveBench}, translating implicit market feedback into constrained numerical reliability updates is more effective than fully relying on the LLM to revise experience reliability scores.

% Overall, the experiments show that, under the current \textsc{FinEvolveBench} setting, structured experience management is better suited to financial sentiment prediction than general-purpose semantic experience systems and reinforcement-learning-based experience management. Formula-based utility updating is also more stable than direct LLM updating. However, the horizon ablation shows that experience retrieval does not improve all prediction windows, and the 5-day results in particular indicate the need for further analysis of retrieval quality and horizon adaptation.

Overall, the experiments support that, under the current \textsc{FinEvolveBench} setting, the effectiveness and stability of structured, formula-updated experience management, while the horizon ablation reveals that its benefits remain conditional on retrieval quality and prediction horizon.

\section{Conclusion}

We present \textbf{T}ree-\textbf{o}f-\textbf{E}xperience (ToE), a structured experience-management framework that aligns experience organization with the hierarchical reasoning process of LLM agents. By representing analytical perspectives as tree nodes and calibrating their reliability through environmental feedback, ToE supports fine-grained outcome attribution, systematic experience updating, perspective-level transfer, and efficient retrieval. Experiments on \textsc{Game of 24} show that ToE improves reasoning accuracy while substantially reducing LLM calls. On \textsc{FinEvolveBench}, where task repetition is low and outcome feedback is delayed and implicit, existing memory mechanisms often provide inconsistent gains or even degrade performance. In contrast, ToE achieves an average \textbf{tsIC} improvement of 41.24 \% across all evaluated settings and remains effective across different backbone LLMs.

\bibliography{main}

\newpage
% \section{Appendix}
\label{sec:appendix}

% =========================================================
% Appendix numbering: Figure A1, Table A1, Equation (A1)
% =========================================================
\setcounter{figure}{0}
\setcounter{table}{0}
\setcounter{equation}{0}

\renewcommand{\thefigure}{A\arabic{figure}}
\renewcommand{\thetable}{A\arabic{table}}
\renewcommand{\theequation}{A\arabic{equation}}

% Appendix-local float settings. Explanatory text remains before the
% corresponding floats, while the final full-width comparison block may
% share a page with the preceding horizon-ablation table.
\setcounter{topnumber}{1}
\setcounter{bottomnumber}{1}
\setcounter{totalnumber}{2}
\setcounter{dbltopnumber}{2}
\renewcommand{\topfraction}{0.85}
\renewcommand{\bottomfraction}{0.75}
\renewcommand{\textfraction}{0.02}
\renewcommand{\floatpagefraction}{0.80}
\renewcommand{\dbltopfraction}{0.98}
\renewcommand{\dblfloatpagefraction}{0.95}

% Keep stacked full-width appendix floats close to the top of the page
% instead of distributing large elastic gaps between them.
\makeatletter
\setlength{\@dblfptop}{0pt}
\setlength{\@dblfpsep}{12pt}
\setlength{\@dblfpbot}{0pt plus 1fil}
\makeatother

\subsection{Structured Experience Representation}
\label{app:structured_experience}

Table~\ref{tab:appendix_experience_example} illustrates the structured
format used to store reusable financial-analysis experiences. Each
experience contains a \textit{factor}, an \textit{analysis direction},
horizon-specific experience-reliability values, and selected maintenance metadata, including recall and hit counts and an experience identifier (EID).

The \textit{factor} and \textit{analysis direction} together form a
reusable reasoning path. The implementation maintains a separate
reliability value for the experience at each of the 10-, 20-, and 40-trading-day horizons.The \textit{recall} counts how often the complete experience path is retrieved as a candidate, rather than referring to the conventional evaluation metric. The \textit{hit} is recorded only when the retrieved path is actually used for reasoning and receives environmental feedback; thus, the hit count does not exceed the recall count. The \textit{eid} is a unique identifier for the complete root-to-leaf path, enabling consistent retrieval, tracking, and reliability updating.

% The \textit{recall}, \textit{hit}, and
% \textit{eid} fields support experience retrieval and subsequent reliability updating.

\begin{table*}[!tbp]
\centering
\footnotesize
\setlength{\tabcolsep}{4pt}
\renewcommand{\arraystretch}{1.25}

\begin{tabular}{@{}
>{\centering\arraybackslash}m{0.12\textwidth}
>{\centering\arraybackslash}m{0.38\textwidth}
>{\centering\arraybackslash}m{0.27\textwidth}
>{\centering\arraybackslash}m{0.09\textwidth}
>{\centering\arraybackslash}m{0.05\textwidth}
@{}}
\toprule
Factor &
Analysis direction &
Selected impact horizons &
Recall/Hit &
EID \\
\midrule

Macroeconomy &
Pricing capacity utilization from aggregate-demand cycles &
\makebox[0.27\textwidth][c]
{[1.6602, 0.5981, 0.9529]} &
705/208 &
1049 \\

Policy planning &
Supply-side administrative constraints and capacity-entry barriers &
\makebox[0.27\textwidth][c]
{[1.1205, 0.3548, 0.2004]} &
1125/602 &
1044 \\

Geopolitics &
Geopolitical uncertainty, policy flexibility of oil-producing
countries, and supply-side capacity disruptions &
\makebox[0.27\textwidth][c]
{[1.4600, 0.7345, 0.8703]} &
22446/13209 &
1040 \\

\bottomrule
\end{tabular}

\caption{Examples of structured experiences in financial sentiment
analysis. For compact presentation, the table displays selected
horizon-specific experience-reliability values.}
\label{tab:appendix_experience_example}
\end{table*}

The examples show how an abstract analytical perspective is retained
independently of any single news item, while its horizon-specific
reliability is updated from delayed market feedback.

\subsection{Experimental Details}
\label{app:experimental_details}

\paragraph{Common experimental setup.}

Our main experiments use \textsc{DeepSeek-V4-Flash} as the primary
backbone model. \textsc{Qwen3.6-35B-A3B} is used as an additional
backbone-model ablation. The backbone models remain frozen during
prediction and experience updating. The context-window size is set to
32,768 tokens.

Unless otherwise specified, each method is independently run three
times, and the reported results are averaged over the three runs.

% \paragraph{Prediction horizons and news input in FinEvolveBench.}
% \paragraph{FinEvolveBench evaluation protocol.}
\paragraph{FinEvolveBench-specific experimental setup.}

The prediction horizons are
\begin{equation}
H=\{10,20,40\},
\end{equation}
corresponding to 10-, 20-, and 40-trading-day predictions.

For a prediction made on date $T$, the model uses a five-day news window
from day $T-4$ to day $T$. Only target-industry news with an importance
score of at least 8 is retained. The same news window and filtering rule
are used for all compared methods.

% \paragraph{Evaluation periods in FinEvolveBench.}

For each prediction horizon, we report results over two temporal
evaluation periods: \textit{Overall} and \textit{Exploitation}. The
\textit{Overall} period covers the complete evaluation interval from
January 2, 2025 to March 31, 2026.

Let $\mathcal{D}=(d_1,\ldots,d_T)$ denote the chronologically ordered
list of trading days shared by all evaluated methods. The first
one-third of these trading days,
$\{d_1,\ldots,d_{\lfloor T/3 \rfloor}\}$, is defined as the
\textit{Cold-Start} period, during which the experience repository is
still being constructed and its reliability estimates have not yet
stabilized. The remaining trading days,
$\{d_{\lfloor T/3 \rfloor+1},\ldots,d_T\}$, constitute the
\textit{Exploitation} period.

The same temporal boundaries are applied to all methods and all
prediction horizons. The exact \textit{Overall} start date, \textit{Overall} end date,
and \textit{Exploitation} start date are explicitly recorded in the released
evaluation configuration to ensure deterministic reproduction of the
reported results.

\paragraph{Hyperparameters And Implementation Details.}

\paragraph{Details in Game of 24.}

In Game of 24, an analytical perspective is instantiated as the current
subproblem state rather than as an independently verbalized conceptual
label. Each state is represented by the canonicalized multiset of its
remaining numbers and serves as a node in the experience tree. For
example, the state
$\{1,1,10,12\}$ may be transformed into
$\{2,10,12\}$ by combining the two numbers 1, or into
$\{1,11,12\}$ by combining 1 and 10. These successor states represent
different local analytical directions for continuing the solution.

Accordingly, operator selection and number grouping are encoded
implicitly by the transition from a parent state to one of its child
states. A root-to-leaf reasoning path is represented as a sequence of
progressively reduced number states, such as a four-number state,
followed by a three-number state and then a two-number state. The
successful solution suffix stored at a node records a previously
validated continuation from the corresponding subproblem state and can
be reused when the same or a sufficiently overlapping state is
encountered again.

In this task-specific implementation, retrieval first considers an
exactly matched canonical number state and may additionally use
number-set overlap to identify candidate reusable continuations.
Every candidate continuation is checked by the deterministic arithmetic
verifier before it is admitted into the reusable experience repository.
Experience expansion therefore adds the successor states along a newly
validated successful trajectory. States that are equivalent after
number canonicalization share the same key and are consolidated into the
same logical node rather than being stored as duplicate branches.

Unlike FinEvolveBench, Game of 24 provides exact and deterministic
environmental feedback. Given a problem instance $x_t$ and a generated
continuation $a_t$, the task-specific reward is defined as
\begin{equation}
r_t
=
\mathbb{I}
\left[
\operatorname{Verify}(a_t,x_t)=\mathrm{true}
\right]
\in\{0,1\},
\end{equation}
where $\operatorname{Verify}(\cdot)$ checks both the arithmetic
correctness of the generated expression and whether the input numbers
are used legally.

The reliability-update operator in ToE is defined according to the
feedback characteristics of the target task. For a general stochastic
environment, a successful trace provides direct evidence that the selected experience path is feasible under the observed state, and its reliability is therefore immediately assigned the corresponding success reward. By contrast, an unsuccessful trace does not necessarily imply that the experience path is intrinsically invalid, since failure may arise from environmental stochasticity, noisy feedback, or imperfect downstream execution. We therefore apply a soft penalty to unsuccessful experiences rather than resetting their reliability to zero.

\begin{equation}
Q_{\mathrm{new}}(e)
=
\begin{cases}
r_t, & \text{if } r_t > 0, \\[4pt]
(1-\eta_{\mathrm{task}})Q_{\mathrm{old}}(e)
+\eta_{\mathrm{task}}r_t, & \text{if } r_t \leq 0.
\end{cases}
\label{eq:asymmetric_reliability_update}
\end{equation}

where $\eta_{\mathrm{task}}\in[0,1]$ controls the strength of the reliability decay following an unsuccessful outcome. A larger $\eta_{\mathrm{task}}$ assigns greater weight to the latest failure, whereas a smaller value preserves more of the historical reliability estimate. This asymmetric design treats success as positive evidence of path feasibility while interpreting failure as probabilistic negative evidence rather than definitive invalidation.
% reliability may be updated incrementally as
% \begin{equation}
% Q_{\mathrm{new}}(e)
% =
% (1-\eta_{\mathrm{task}})Q_{\mathrm{old}}(e)
% +
% \eta_{\mathrm{task}}r_t,
% \end{equation}
% where $\eta_{\mathrm{task}}$ controls how rapidly newly observed
% environmental evidence replaces the previous reliability estimate.

In Game of 24, the outcome of a fixed state--continuation pair is
deterministic: a mathematically valid continuation remains valid
whenever the same canonical subproblem state is encountered, whereas an
invalid continuation remains invalid. Consequently, repeated
observations are not required to estimate an uncertain success
probability. We therefore set
$\eta_{\mathrm{G24}}=1$, under which the reliability update collapses to
\begin{equation}
Q_{\mathrm{new}}(e)
=
r_t
\in\{0,1\}.
\end{equation}
Equivalently,
\begin{equation}
\operatorname{UpdateQ}_{\mathrm{G24}}
\bigl(Q_{\mathrm{old}}(e),r_t\bigr)
=
r_t.
\end{equation}

A verifier-confirmed continuation receives $Q(e)=1$ and is admitted
into the reusable experience repository. A continuation that fails
verification receives $Q(e)=0$ and is immediately rejected rather than
being stored as a reusable experience. Thus, the absence of persistent
failed continuations is a consequence of the binary reliability
criterion and the repository admission rule, rather than the absence of
environmental feedback or reliability updating.

This task-specific binary update differs from the reliability update
used in FinEvolveBench. Financial outcomes are delayed, noisy, and
continuous, so a single observation cannot determine whether an
analytical experience is intrinsically reliable. FinEvolveBench
therefore retains graded reliability values and updates them
incrementally across repeated environmental observations. In contrast,
the exact verifier in Game of 24 causes the general task-dependent
reliability update to collapse into a binary validity mapping after one
observation.

Accordingly, Game of 24 implements a deterministic, state-based
specialization of ToE. Analytical perspectives, hierarchical reasoning
paths, environmental calibration, expansion, and consolidation are
instantiated through canonical subproblem states, state transitions,
verifier-confirmed solution suffixes, and binary correctness feedback.

% We use the follow hyperparameter settings across all experiments in FinEvolveBench. The
% experience-tree depth is set to $L=3$. The number of candidates retained
% at the first level is set to $k_1=8$, and the cosine-similarity threshold
% for leaf-level retrieval is set to $\lambda=0.8$.

% The initial experience-reliability value is set to $q_0=1$, and the
% learning rate for formula-based reliability updating is set to
% $\eta=10$. Each prediction horizon maintains a separate reliability
% value. Newly created experiences are initialized with reliability 1,
% and the reliability values are restricted to $[0,2]$.

% At most one new analysis direction is proposed in each adaptive
% expansion operation. The generated sentiment scores are restricted to
% $[-1,1]$, and confidence scores are restricted to $[0,1]$.

\paragraph{Hyperparameters in FinEvolveBench.}
We use the follow hyperparameter settings across all experiments in FinEvolveBench. The
experience-tree depth is set to $L=3$. The number of candidates retained
at the first level is set to $k_1=8$, and the cosine-similarity threshold
for leaf-level retrieval is set to $\lambda=0.8$.

The initial experience-reliability value is set to $q_0=1$, and the
learning rate for formula-based reliability updating is set to
$\eta=10$. Each prediction horizon maintains a separate reliability
value. Newly created experiences are initialized with reliability 1,
and the reliability values are restricted to $[0,2]$.

At most one new analysis direction is proposed in each adaptive
expansion operation. The generated sentiment scores are restricted to
$[-1,1]$, and confidence scores are restricted to $[0,1]$.

\paragraph{Tree Representation and Decoupled Execution in FinEvolveBench.}
For \textsc{FinEvolveBench}, the Tree-of-Experience is organized into three hierarchical levels: industry, factor, and analysis\_direction. We adopt a path-based tabular representation in which each row corresponds to a complete root-to-leaf path. Rows sharing the same industry value form an industry-level subtree, while rows sharing the same (industry, factor) prefix form a factor-level subtree. The resulting CSV representation preserves the logical tree structure through shared hierarchical prefixes, while supporting efficient subtree filtering, path insertion, and reliability updates without explicit parent pointers or nested node objects.

Routing through the tree constitutes a hierarchical reasoning process conditioned on the news content. The model first infers the industry to which the news is most relevant. Conditioned on both the news and the inferred industry, it then identifies the factor that best characterizes the underlying channel of influence. Finally, given the resulting (industry, factor) prefix, the model retrieves and selects analysis directions from the corresponding subtree. The three levels therefore form a logically dependent reasoning path rather than a set of independently observed labels.

To accelerate large-scale experimentation, we decouple the execution of the first two reasoning stages from direction-level retrieval and final prediction. Specifically, the industry and factor decisions are inferred in advance from each news item and stored as an intermediate routing prefix. During the subsequent prediction stage, this cached prefix is used to locate the corresponding factor-level subtree, from which the model continues the reasoning process by selecting relevant analysis\_direction nodes and incorporating their reliability states.

This decoupled implementation does not remove or bypass the hierarchical reasoning involved in industry and factor selection. It only avoids repeatedly executing the complete three-stage routing chain within every prediction query. The logical dependency from news content to industry, from industry to factor, and from factor to analysis direction remains unchanged; only the execution of the earlier reasoning stages is separated from the downstream prediction call and reused across experiments.

\paragraph{Formula-based experience-reliability update in
FinEvolveBench.}
% Let analytical path $\mathbf{\Pi} = (p^{(1)}, \dots, p^{(L)})$,
% Let $\mathbf{P}^{*}$ denote the selected experience path. In
In
FinEvolveBench, $n_{\mathrm{hit}}$ is defined as the historical
\emph{news-level retrieval count} of that experience. Experience
reranking is performed independently for each retained news item, and
$n_{\mathrm{hit}}$ is incremented once whenever the experience is
selected for one news item. Therefore, if the same experience is
selected for five news items on the same prediction date, its hit count
increases by five, even though these news items are subsequently
aggregated into one industry-level prediction.

Accordingly, $n_{\mathrm{hit}}$ measures accumulated retrieval exposure
at the news-item level. It does not denote the number of aggregated
industry-level predictions, the number of pending horizon-specific
update records, or the number of times that delayed environmental
feedback becomes available. The reliability
of the selected experience path is updated as
\begin{equation}
\mathbf{Q}_{\mathrm{new}}(e)
\leftarrow
\mathbf{Q}_{\mathrm{old}}(e)
+
\eta
\frac{
1-
\left(
\frac{
\mathbf{Q}_{\mathrm{old}}(e)-q_0
}{
q_0
}
\right)^2
}{
1+\log(1+n_{\mathrm{hit}})
}
\odot
\mathbf{r}_{\mathrm{env}}.
\end{equation}

After updating, the experience-reliability values are clipped to
$[0,2]$. The usage count reduces the update size for frequently used
paths, while the boundary term prevents reliability values from changing
too rapidly near the lower and upper limits.

\paragraph{Fairness controls.}

All methods use the same backbone model, context-window size,
news-processing pipeline, news window, importance-score threshold,
prediction prompt, evaluation order, prediction horizons, feedback
definition, and evaluation metrics.

The experience-based methods use the same pre-test historical tasks for
experience initialization. They also follow the same temporal
constraints. Therefore, the main differences among these methods come
from experience representation, retrieval, and updating.

\paragraph{Data availability and reproducibility.}

The supplementary material does not include the complete raw financial
news corpus and market-price files because these data contain a large
number of long-form news documents and high-frequency temporal records,
whose total storage size substantially exceeds the supplementary
material size limit.

Instead, the released package provides the complete method-specific
implementation, including news-processing interfaces, hierarchical
experience construction and storage, experience retrieval and merging,
horizon-specific prediction, delayed reliability updating, and the
evaluation scripts used to compute csIC and tsIC. It also specifies the
required directory structure, input fields, prediction formats,
evaluation horizons, temporal splits, and configuration parameters.

Researchers can reproduce the reported experimental procedure by
organizing the corresponding benchmark data according to the documented
input structure and running the supplied scripts and configurations.
No additional undisclosed method-specific processing or evaluation
procedure is required to reproduce the proposed method.

\paragraph{Compared methods.}

We compare five main methods in RQ1 and RQ2. Except for Baseline, all
methods use the same structured news annotations and prediction prompt.

\begin{itemize}

  \item \textbf{Baseline}: Following previous financial market
  prediction studies
  ~\citep{wang2018combining,mohan2019stock,
  lopezlira2023chatgpt,wang2024mananet}, the LLM predicts a discrete
  sentiment value in $\{-1,0,1\}$ for each retained news item. The
  importance-weighted average of the news sentiments from $T-4$ to $T$
  is used as the industry sentiment prediction. This method does not use
  experience retrieval.

  \item \textbf{Pipe}: The pipeline generates structured annotations,
  including importance scores and related industries. It aggregates
  target-industry news from $T-4$ to $T$ with an importance score of at
  least 8 and sends the news to the LLM prediction module. This method
  does not use experience retrieval.

  \item \textbf{Pipe+mem0}: This method uses
  mem0~\citep{chhikara2025mem0} for experience storage and semantic
  retrieval. The retrieved experiences and aggregated news are passed
  together to the prediction module.

  \item \textbf{Pipe+MemRL}: This method uses
  MemRL~\citep{memrl2026} for experience retrieval and updating. The
  retrieved experiences are used during prediction, and the memory is
  updated after delayed feedback becomes available.

  \item \textbf{Pipe+ToE (ours)}: This method uses a depth-3
  Tree-of-Experience. It retrieves an experience path through
  internal-level alignment and leaf-level adaptation. The selected
  experience and aggregated news are passed to the prediction module.
  Experience-reliability values are updated after delayed feedback becomes
  observable.

\end{itemize}

% \paragraph{Update-strategy ablation.}

% For RQ3, we compare two experience-reliability update strategies under
% the same retrieval and prediction pipeline. \textbf{Pipe+formula} uses
% the formula-based update described above. \textbf{Pipe+llm} asks the
% frozen LLM to update the reliability of the experience after receiving
% delayed feedback.

% The two methods use the same experience retrieval procedure, prediction
% prompt, backbone model, and evaluation metrics. Therefore, this
% comparison isolates the effect of the experience-reliability update
% rule.

\paragraph{Evaluation metrics.}

We use the information coefficient (IC) as the main evaluation metric.
The task produces continuous sentiment factors rather than categorical
labels, so the predicted values are evaluated according to their
correlation with future market returns.

\textbf{Cross-sectional IC (csIC)} measures the correlation between
predicted sentiment values and future excess returns across industries
on the same trading day:
\begin{equation}
\operatorname{csIC}_{t,h}
=
\operatorname{Corr}_{i}
\left(
s_{i,t,h},
\alpha_{i,t,h}
\right).
\end{equation}

The final csIC for horizon $h$ is
\begin{equation}
\operatorname{csIC}_{h}
=
\frac{1}{T}
\sum_{t=1}^{T}
\operatorname{csIC}_{t,h}.
\end{equation}

\textbf{Time-series IC (tsIC)} measures the correlation between
predicted sentiment values and future excess returns across time for
the same industry:
\begin{equation}
\operatorname{tsIC}_{i,h}
=
\operatorname{Corr}_{t}
\left(
s_{i,t,h},
\alpha_{i,t,h}
\right).
\end{equation}

The final tsIC for horizon $h$ is
\begin{equation}
\operatorname{tsIC}_{h}
=
\frac{1}{N}
\sum_{i=1}^{N}
\operatorname{tsIC}_{i,h}.
\end{equation}

\subsection{Cross-Sectional Performance During Evolution}
\label{app:csic_evolution}

Table~\ref{tab:csic_evolution} reports the evolution-stage csIC of the
four pipeline-based methods over the 10-, 20-, and 40-trading-day
horizons. Pipe serves as the non-experience reference, while the other
three methods augment the same prediction pipeline with different
experience-management mechanisms. This comparison therefore focuses on
whether experience retrieval and evolution improve cross-sectional
ranking beyond the structured prediction pipeline itself.

Pipe+ToE achieves the highest csIC in all six backbone--horizon
settings. With \textsc{DeepSeek-V4-Flash}, its csIC values are 0.0873,
0.1210, and 0.1371 for the 10-, 20-, and 40-day horizons, respectively.
With \textsc{Qwen3.6-35B-A3B}, it obtains 0.0834, 0.1221, and 0.1055.
The consistent advantage across both backbone models indicates that the
hierarchical organization and reliability-aware evolution of ToE
provide more robust medium- and long-horizon experience augmentation
than flat semantic retrieval or alternative memory updating.

\begingroup
\centering
\setlength{\tabcolsep}{2.2pt}
\renewcommand{\arraystretch}{1.12}

\resizebox{\linewidth}{!}{%
\begin{tabular}{@{}lcccccc@{}}
    \toprule

    & \multicolumn{3}{c}{DeepSeek-V4-Flash}
    & \multicolumn{3}{c}{Qwen3.6-35B-A3B} \\

    \cmidrule(lr){2-4}
    \cmidrule(lr){5-7}

    \textbf{Method}
    & 10d & 20d & 40d
    & 10d & 20d & 40d \\

    \midrule

    Pipe
    & 0.0534 & 0.0837 & 0.0589
    & 0.0456 & 0.0600 & 0.0674 \\

    Pipe+mem0
    & 0.0524 & 0.0623 & 0.0682
    & 0.0413 & 0.0525 & 0.0611 \\

    Pipe+MemRL
    & 0.0542 & 0.0344 & 0.0356
    & 0.0219 & 0.0467 & 0.0064 \\

    Pipe+ToE
    & \textbf{0.0873}
    & \textbf{0.1210}
    & \textbf{0.1371}
    & \textbf{0.0834}
    & \textbf{0.1221}
    & \textbf{0.1055} \\

    \bottomrule
\end{tabular}%
}

\endgroup
\captionof{table}{Exploitation-stage csIC across the 10-, 20-, and
40-trading-day horizons.}
\label{tab:csic_evolution}

\subsection{Prompt Templates}
\label{app:prompt_templates}

For reproducibility, we present the three LLM prompt templates used in
our implementation: experience reranking and adaptive expansion,
experience merging, and experience-conditioned sentiment prediction.
Runtime inputs are represented by placeholders enclosed in braces.

\paragraph{Experience reranking and adaptive expansion.}

The reranking prompt selects historical analytical perspectives that
are applicable to the current news. If no suitable experience exists,
the prompt may propose one new analysis direction.

\begin{table*}[!tbp]
\centering

\begin{tcolorbox}[
  enhanced,
  width=0.94\textwidth,
  colback=white,
  colframe=black!65,
  boxrule=0.7pt,
  arc=2pt,
  outer arc=2pt,
  left=10pt,
  right=10pt,
  top=7pt,
  bottom=7pt,
  colbacktitle=black!72,
  coltitle=white,
  fonttitle=\bfseries\footnotesize,
  title={Example Prompt for Experience Reranking and Adaptive Expansion},
  titlerule=0pt,
  before skip=0pt,
  after skip=0pt
]

\footnotesize
\setlength{\parindent}{0pt}
\setlength{\parskip}{2.5pt}

\textbf{ROLE}

You are a senior quantitative researcher and experience-ranking expert,
focusing on selecting the analytical perspectives from historical
analysis experiences that best match the current news event.

\textbf{TASK}

You will receive:

\textbf{1.} One or more \textbf{news items}, including their dates and
importance.

\textbf{2.} A set of \textbf{historical analytical experiences}, each
containing \texttt{factor}, \texttt{analysis\_direction}, and impact
scores for different horizons.

Determine whether the analytical perspective of each historical
experience is applicable to sentiment judgment for the current news,
and output the \texttt{eid}s of the top-$K$ most relevant experiences
in descending order of relevance.

\textbf{PROPOSING A NEW ANALYSIS DIRECTION}

If the analytical perspectives in all historical experiences are
unsuitable for the current news, you may propose a new analysis
direction. The new direction must satisfy the following constraints:

\textbf{1. Format requirement.}

\texttt{analysis\_direction} must use the format
``From the perspective of [an intrinsic property of the factor].''

Example: ``From the perspective of how the discount rate anchors the
valuation of capital-intensive assets.''

The \texttt{analysis\_direction} must describe an intrinsic-property
dimension or pricing-interpretation perspective through which the
factor acts on the target industry.

\textbf{2. Conclusion-free expression.}

Provide only a logical perspective for reasoning. Do not include
directional conclusions such as price increases, price decreases,
bullish effects, or bearish effects.

\textbf{3. Entity-free expression.}

Do not include specific assets, countries, institutions, or exact
indicator values. Use abstract industrial and financial terminology.

\textbf{4. Impact horizon.}

The \texttt{impact\_horizon} fields represent the reliability of the
experience over different time windows. Values must
lie in $[0.0,2.0]$. When uncertain, use the initial value 1.

Even when a new analysis direction is proposed,
\texttt{ranked\_eids} must still be returned and may be an empty list.
Only one new perspective may be proposed each time.

\textbf{RUNTIME INPUT}

\texttt{\{news\_content\}}

\texttt{\{historical\_analytical\_experiences\}}

\textbf{OUTPUT REQUIREMENT}

Return a single JSON object directly. Do not include additional
explanatory text or Markdown code-block markers.

\begin{lstlisting}[
  basicstyle=\ttfamily\scriptsize,
  columns=fullflexible,
  keepspaces=true,
  showstringspaces=false,
  breaklines=true,
  frame=none,
  aboveskip=2pt,
  belowskip=2pt
]
{
  "ranked_eids": [eid1, eid2, ...],
  "proposed_new_directions": [
    {
      "analysis_direction":
        "From the perspective of [an intrinsic-property dimension of the factor]",
      "impact_horizon_10d": 0.0,
      "impact_horizon_20d": 0.0,
      "impact_horizon_40d": 0.0
    }
  ]
}
\end{lstlisting}

Here, \texttt{ranked\_eids} contains only experiences that are applicable
to the current task. \texttt{proposed\_new\_directions} is used only
when no existing experience is suitable.

\end{tcolorbox}

\caption{Example prompt for experience reranking and adaptive
expansion.}
\label{tab:rerank_prompt}
\end{table*}

\paragraph{Experience merging.}

When a newly proposed direction is highly similar to an existing
experience, the merging prompt determines whether the two experiences
should be combined.

\begin{table*}[!tbp]
\centering

\begin{tcolorbox}[
  enhanced,
  width=0.94\textwidth,
  colback=white,
  colframe=black!65,
  boxrule=0.7pt,
  arc=2pt,
  outer arc=2pt,
  left=10pt,
  right=10pt,
  top=7pt,
  bottom=7pt,
  colbacktitle=black!72,
  coltitle=white,
  fonttitle=\bfseries\footnotesize,
  title={Example Prompt for Experience Merging},
  titlerule=0pt,
  before skip=0pt,
  after skip=0pt
]

\footnotesize
\setlength{\parindent}{0pt}
\setlength{\parskip}{2.5pt}

\textbf{ROLE}

You are an experience-quality reviewer responsible for determining
whether a newly proposed experience should be merged with an existing
experience.

\textbf{CORE PRINCIPLES}

\textbf{1. Analysis-direction format.}

\texttt{analysis\_direction} must use the format
``From the perspective of [an intrinsic property or logical exposure
of the factor]'' and must not contain a directional market conclusion.

\textbf{2. Experience reliability.}

Each \texttt{impact\_horizon\_xd} field represents the reliability of
the experience over the corresponding time window. Values must lie in
$[0.0,2.0]$.

\textbf{TASK}

The system intends to add a new experience but has identified a highly
similar existing experience in the same industry context. Decide whether
the new experience should be merged with the existing experience or
retained independently.

\textbf{DECISION CRITERIA}

\textbf{Merge} when the two analysis directions describe the same or
highly complementary analytical dimension.

\textbf{Do not merge} when they describe different analytical
dimensions and retain independent value.

\textbf{RUNTIME INPUT}

\begin{lstlisting}[
  basicstyle=\ttfamily\scriptsize,
  columns=fullflexible,
  keepspaces=true,
  showstringspaces=false,
  breaklines=true,
  frame=none,
  aboveskip=2pt,
  belowskip=2pt
]
Industry: {l1name}

New experience:
- analysis_direction: {new_analysis_direction}
- factor: {new_factor}
- impact_horizon_10d: {new_impact_10d}
- impact_horizon_20d: {new_impact_20d}
- impact_horizon_40d: {new_impact_40d}

Similar existing experience:
- eid: {existing_eid}
- analysis_direction: {existing_analysis_direction}
- factor: {existing_factor}
- impact_horizon_10d: {existing_impact_10d}
- impact_horizon_20d: {existing_impact_20d}
- impact_horizon_40d: {existing_impact_40d}
- life: {existing_life}
\end{lstlisting}

\textbf{OUTPUT REQUIREMENT}

Return a single JSON object directly. Do not include additional
explanatory text or Markdown code-block markers. The \texttt{reason}
must be concise.

\begin{lstlisting}[
  basicstyle=\ttfamily\scriptsize,
  columns=fullflexible,
  keepspaces=true,
  showstringspaces=false,
  breaklines=true,
  frame=none,
  aboveskip=2pt,
  belowskip=2pt
]
{
  "should_merge": true | false,
  "reason": "A concise justification for the decision",
  "merged_analysis_direction":
    "From the perspective of [a unified intrinsic-property dimension]",
  "merged_impact_horizon_10d": 0.0,
  "merged_impact_horizon_20d": 0.0,
  "merged_impact_horizon_40d": 0.0
}
\end{lstlisting}

The merged analysis direction must retain the required format, and all
merged horizon-specific reliability values must remain in $[0.0,2.0]$.

\end{tcolorbox}

\caption{Example prompt for experience merging.}
\label{tab:merge_prompt}
\end{table*}

\paragraph{Experience-conditioned sentiment prediction.}

The prediction prompt combines the selected experiences with the
target-industry news. It outputs a reasoning chain, a key factor,
horizon-specific sentiment scores, and a confidence score.

\begin{table*}[!tbp]
\centering

\begin{tcolorbox}[
  enhanced,
  width=0.94\textwidth,
  colback=white,
  colframe=black!65,
  boxrule=0.7pt,
  arc=2pt,
  outer arc=2pt,
  left=10pt,
  right=10pt,
  top=7pt,
  bottom=7pt,
  colbacktitle=black!72,
  coltitle=white,
  fonttitle=\bfseries\footnotesize,
  title={Example Prompt for Experience-Conditioned Sentiment Prediction},
  titlerule=0pt,
  before skip=0pt,
  after skip=0pt
]

\footnotesize
\setlength{\parindent}{0pt}
\setlength{\parskip}{2.5pt}

\textbf{ROLE}

You are a senior quantitative researcher specializing in event-driven
strategies and industry-factor modeling for the Chinese A-share market.
You are skilled at converting unstructured financial news into
quantitative signals.

\textbf{TASK}

Analyze the news summary for the given industry and assess the event's
sentiment direction and logical transmission strength over the next
1, 5, 10, 20, and 40 trading days.

\textbf{HISTORICAL EXPERIENCE GUIDANCE}

If historical analytical experiences are provided before the news in
the form ``[From the perspective of ...],'' use their
\texttt{analysis\_direction}s and horizon-specific impact values when
constructing the \texttt{logic\_chain}.

Each impact value lies in $[0.0,2.0]$ and represents the reliability of
the experience over the corresponding horizon. It has no directional
meaning and does not itself indicate positive or negative sentiment.

\textbf{RUNTIME INPUT}

\texttt{\{target\_industry\}}

\texttt{\{historical\_analytical\_experiences\}}

\texttt{\{news\_summary\}}

\textbf{OUTPUT REQUIREMENT}

Return a single JSON object directly. Do not include additional
explanatory text or Markdown code-block markers. The JSON object must
contain:

\textbf{1.} \texttt{logic\_chain}: a causal analysis of how the news is
transmitted to the target industry.

\textbf{2.} \texttt{key\_factor}: the most important driver, summarized
in no more than 10 words.

\textbf{3.} \texttt{sentiment\_score}: continuous scores in
$[-1.0,1.0]$ for all five horizons.

\textbf{4.} \texttt{confidence\_score}: confidence in the analysis,
with a value in $[0.0,1.0]$.

\begin{lstlisting}[
  basicstyle=\ttfamily\scriptsize,
  columns=fullflexible,
  keepspaces=true,
  showstringspaces=false,
  breaklines=true,
  frame=none,
  aboveskip=2pt,
  belowskip=2pt
]
{
  "logic_chain": "A detailed causal transmission chain",
  "key_factor": "The dominant driver",
  "sentiment_score": {
    "horizon_10d": 0.0,
    "horizon_20d": 0.0,
    "horizon_40d": 0.0
  },
  "confidence_score": 0.0
}
\end{lstlisting}

\end{tcolorbox}

\caption{Example prompt for experience-conditioned, horizon-specific
sentiment prediction.}
\label{tab:prediction_prompt}
\end{table*}

\end{document}